\documentclass{article}

\usepackage{PRIMEarxiv}

\usepackage[utf8]{inputenc}
\usepackage[T1]{fontenc}
\usepackage{xcolor}
\newcommand{\aqrevision}[1]{#1}
\usepackage{url}
\usepackage{booktabs}
\usepackage{amsfonts}
\usepackage{microtype}
\usepackage{fancyhdr}
\usepackage{graphicx}

\usepackage{amsmath,amssymb}
\usepackage{multirow}
\usepackage{makecell}
\usepackage[labelsep=period]{caption}
\newcommand{\sctablesetup}{%
    \footnotesize
    \setlength{\tabcolsep}{4pt}%
    \renewcommand{\arraystretch}{1.12}%
}
\usepackage{placeins}
\usepackage{hyperref}
\usepackage{originalrefs}

\let\citep\cite
\let\citet\cite

\title{A swap-adversarial framework for domain generalization in pre- and post-stimulation Parkinsonian state classification}

\author{
  Seongwon Jin\textsuperscript{1}, Hanseul Choi\textsuperscript{1}, Gaeun Kim\textsuperscript{2}, \\
  \textbf{Sunggu Yang\textsuperscript{2,3}, Sungho Park\textsuperscript{4,*}, Jibum Kim\textsuperscript{1,3,*}} \\[1ex]
  \textsuperscript{1}Department of Computer Science and Engineering, \\
  Incheon National University, Yeonsu-gu, 22012, Incheon, Republic of Korea \\
  \textsuperscript{2}Department of Nanobioengineering, \\
  Incheon National University, Yeonsu-gu, 22012, Incheon, Republic of Korea \\
  \textsuperscript{3}Center for Brain-Machine Interface, \\
  Incheon National University, Yeonsu-gu, 22012, Incheon, Republic of Korea \\
  \textsuperscript{4}Department of Artificial Intelligence, \\
  Inha University, Michuhol-gu, 22212, Incheon, Republic of Korea \\[1ex]
  \textsuperscript{*}Correspondence to: Sungho Park (\texttt{sungho.park@inha.ac.kr}) \\
  or Jibum Kim (\texttt{jibumkim@inu.ac.kr})
}

\hypersetup{
  colorlinks=true,
  linkcolor=red,
  citecolor=green,
  urlcolor=black,
  pdfborder={0 0 0},
  pdftitle={A swap-adversarial framework for domain generalization in pre- and post-stimulation Parkinsonian state classification},
  pdfauthor={Seongwon Jin, Hanseul Choi, Gaeun Kim, Sunggu Yang, Sungho Park, Jibum Kim},
  pdfkeywords={Electrocorticography, Electroencephalography, Classification, Domain generalization, Deep learning}
}

\begin{document}
\maketitle

\begin{abstract}
We propose a novel swap-adversarial framework that mitigates high inter-subject variability and the high-dimensional low-sample-size problem in electrocorticography (ECoG) data. It achieves robust domain generalization across ECoG and electroencephalography (EEG)-based brain-computer interface datasets. Our framework integrates (1) robust preprocessing, (2) inter-subject balanced channel swap (ISBCS) for cross-subject augmentation, and (3) domain-adversarial learning (DAL) to suppress subject-specific bias. The ISBCS method is a bio-inspired channel swapping strategy that exchanges only spatially corresponding channels across subjects, guided by a motor-cortex brain map, to mitigate inter-subject distribution differences. The DAL strategy encourages the model to learn task-relevant shared features. In addition, we introduce a new ECoG dataset, the first reproducible benchmark for classifying neural states before and after electrical stimulation, constructed from long-term ECoG recordings of 6-hydroxydopamine-induced PD rat models. We validate the effectiveness of this framework through cross-subject and cross-cohort evaluations on ECoG data and cross-dataset evaluations on public EEG data. Our results show that the proposed framework exhibits the best and robust generalization performance both for our newly constructed ECoG and public EEG datasets.
\end{abstract}

\keywords{Electrocorticography \and Electroencephalography \and Classification \and Domain generalization \and Deep learning}

\section{Introduction}
\label{sec1}
Parkinson's disease (PD) is a progressive neurodegenerative movement disorder characterized by bradykinesia, rigidity, tremor, and postural instability \citep{kalia2015parkinson}. Early detection and treatment remain challenging, as clinical diagnosis is generally possible only after the onset of visible motor symptoms \citep{obeso2017past, dorsey2018global}. While therapeutic interventions such as deep brain stimulation can alleviate motor symptoms, their application is fundamentally restricted to clinically manifest stages \citep{benabid2009deep}.
Since PD-related neurophysiological changes may be reflected in brain signals before motor symptoms become apparent, recent studies have combined brain signals with deep learning models to enable early diagnosis \citep{chaturvedi2017quantitative, oh2020deep, abumalloh2024parkinson, neves2024parkinson}.
Among brain recording modalities, Electroencephalography (EEG) has been widely used owing to its non-invasive nature, but its limited spatial resolution and attenuation of high-frequency components make it difficult to capture characteristic PD patterns such as local $\beta$-band fluctuations and pathological high-frequency oscillations \citep{ali2022predictive, kim2025explainable, lachaux2012high, srinivasan2007eeg}. Electrocorticography (ECoG), recorded directly from the cortical surface, offers higher spatial resolution, signal-to-noise ratio (SNR), and frequency bandwidth, making it a promising alternative \citep{kanth2020electrocorticogram}.

However, invasive ECoG studies involving human subjects are limited by ethical and safety concerns, and most prior work has relied on animal models of PD induced by 6-hydroxydopamine (6-OHDA) \citep{kim2025explainable, shin2025wireless, wang2024longitudinal}. Although a publicly available ECoG dataset from 6-OHDA-induced rats exists \citep{wang2024longitudinal}, its inconsistent channel configurations across subjects and recording states make it unsuitable for training and the reliable evaluation of deep learning models. To fill this gap, we constructed and {released RATCOP (Rat ECoG dataset for Parkinson's disease)}, the first ECoG-based benchmark for classifying neural states before and after electrical stimulation, based on long-term recordings from 6-OHDA-lesioned rats \citep{shin2025wireless}, with standardized data partitioning and cross-validation protocols for unseen-domain evaluation.

A key challenge in ECoG-based classification is the substantial inter-subject variability arising from anatomical differences in electrode placement and cortical folding, which is further amplified by temporal non-stationarity during long-term recordings \citep{miller2010cortical, ung2017intracranial}. This variability causes subject dependency, where models overfit to specific subjects and degrade on unseen ones \citep{ali2022predictive}. Domain generalization approaches, particularly adversarial learning \citep{ganin2016domain}, can mitigate this by learning subject-invariant representations, yet their application to ECoG has remained unexplored. Moreover, the high-dimensional and low-sample-size (HDLSS) nature of ECoG data leads to unstable adversarial training \citep{sliwowski2023impact}, and the strong entanglement between subject-specific and task-relevant features risks unintended removal of diagnostically meaningful information \citep{haufe2018elucidating, memar2025rise, dayanik2021disentangling}.

To address these challenges, we present the swap-adversarial framework (SAF), which integrates three components in a sequential pipeline. First, artifact subspace reconstruction (ASR) \citep{mullen2015real} removes high-amplitude artifacts from raw ECoG recordings, ensuring that subsequent stages focus on meaningful neurophysiological patterns.
Second, we propose inter-subject balanced channel swap (ISBCS), a bio-inspired augmentation method that swaps only spatially corresponding channels across subjects of the same class based on a brain map \citep{shin2025wireless, leergaard2000somatotopic, penfield1937somatic}. Unlike conventional within-subject augmentations e.g., Mixup \citep{tang2025sdc}) or random channel shuffling \citep{saeed2021learning}, which either fail to reduce inter-subject distributional gaps or disrupt spatial-functional correspondence, ISBCS performs cross-subject, brain-map-guided, and strictly channel-wise exchange. This strategic approach effectively bridges inter-subject distributions while preserving temporal dynamics and class-discriminative information. Third, domain-adversarial learning (DAL) further suppresses any remaining subject-specific features from the representations already de-biased by ISBCS, thereby minimizing the loss of task-relevant information and improving both training stability and generalization.

In the experimental section, we extensively validated the effectiveness of the proposed method from multiple perspectives. First, in cross-subject experiments using ECoG datasets demonstrating its superior performance. In addition, cross-cohort experiments between the wired and wireless ECoG datasets evaluated generalization under a compound domain shift that includes animal-cohort differences and acquisition-related factors. Finally, in cross-dataset evaluations between public EEG benchmarks, the proposed method was effectively applied to the EEG domain and mitigated dataset-level domain gaps, demonstrating strong generalization capability across modalities.

Our main contributions are summarized as follows:
\begin{enumerate}
    \item We propose a novel SAF designed to learn domain-invariant representations. It generates counterfactual samples that swap only spatially corresponding channels across subjects and mitigates inter-subject distribution differences based on a motor-cortex brain map. DAL is subsequently applied to remove the remaining subject-dependent features, thereby enhancing generalization performance while preserving task-relevant information.
    \item We introduce {RATCOP}, a newly constructed and publicly released benchmark dataset comprising ECoG recordings from 6-OHDA-induced PD rat models. This dataset, which includes precisely annotated state labels, will be made available to the research community upon publication to ensure reproducibility.
    \item Through extensive experiments, we demonstrate the robust domain generalization performance of the proposed framework across all experimental settings and benchmark datasets.
\end{enumerate}

\section{Related work}
\label{sec2}
\subsection{ECoG decoding}
\label{sec2.1}
Early studies on PD classification relied on statistical approaches such as spectral and wavelet-based analyses \citep{agrawal2025quantitative, jeong2016wavelet}, which are limited in capturing the nonlinear dynamics underlying PD \citep{obayya2023novel, schlogl1996single}. Subsequent deep learning methods---\citet{oh2020deep} with a convolutional neural network (CNN) and \citet{shah2020dynamical} with the CNN--long short-term memory (LSTM) hybrid DGHNet---improved EEG-based PD classification but remain dependent on subject-specific characteristics, resulting in poor cross-subject generalization. ECoG provides higher spatial resolution, SNR, and frequency bandwidth than EEG \citep{kanth2020electrocorticogram}, yet ECoG-based research remains scarce due to the difficulty of data acquisition. Among existing studies, \citet{bore2020prediction} and \citet{lam2024self} achieved strong within-subject performance for PD and seizure classification but did not evaluate cross-subject generalization. \citet{ji2024bi} and \citet{cui2022cluster} did evaluate cross-subject transfer. However, the datasets used in these studies are not publicly available.
\subsection{Domain generalization}
\label{sec2.2}
Domain generalization aims to learn models that generalize to unseen target domains without accessing target data during training \citep{ganin2016domain, li2018domain, dayal2023madg, shankar2018generalizing, muandet2013domain}. Existing approaches can be categorized into three groups. Distribution alignment methods \citep{bore2020prediction, dayal2023madg} minimize cross-domain feature discrepancies via maximum mean discrepancy (MMD)-based strategies but are insufficient when inter-subject variability is entangled with high-dimensional features, as in ECoG \citep{miller2010cortical}. Invariance and robustness approaches such as invariant risk minimization (IRM) \citep{arjovsky2020invariant} and Group DRO \citep{sagawa2019distributionally} assume well-defined group structures or balanced domain coverage---conditions rarely met with few ECoG subjects. Adversarial learning methods \citep{ganin2016domain, li2018domain, yang2020towards} train a feature encoder against a domain classifier, directly suppressing domain-specific information.

In EEG-based brain-computer interface (BCI), adversarial domain generalization has been successfully combined with complementary techniques: \citet{ma2019reducing} extended domain-adversarial neural network (DANN) to the domain generalization setting for emotion classification, \citet{wang2024dmmr} incorporated Mixup-based latent-space augmentation across subjects, and \citet{tao2024local} modeled subdomain correlations via a low-rank constraint. Among these paradigms, adversarial learning is well suited for ECoG because it directly removes subject-specific information from high-dimensional feature spaces. However, domain generalization has not yet been applied to ECoG, mainly due to the limited availability of datasets.

\begin{figure}[!t]
    \centering
    \includegraphics[width=0.63\linewidth]{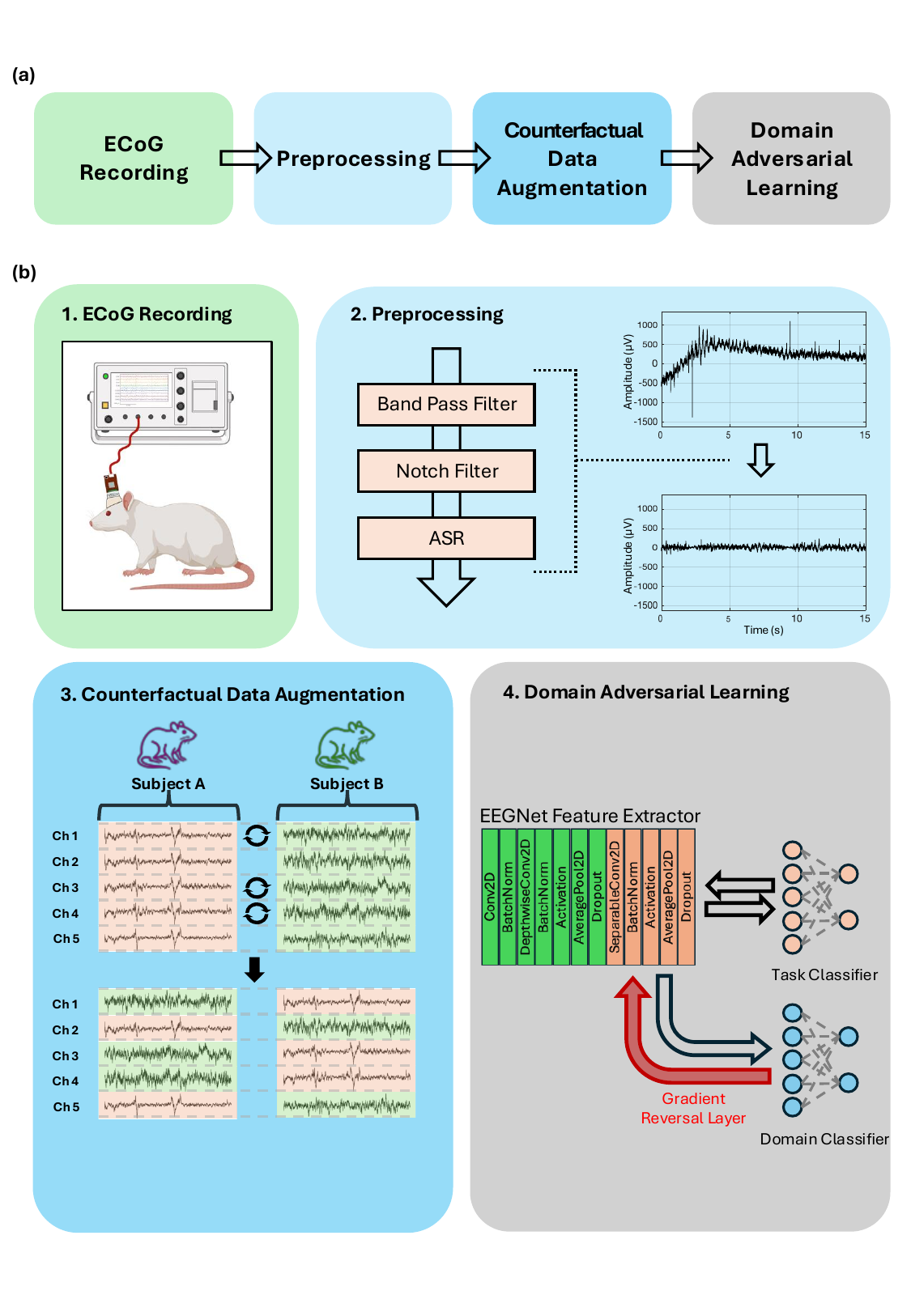}
    \caption{The overall architecture of the SAF. (a) The overall pipeline consists of four main stages: ECoG recording, preprocessing, counterfactual data augmentation, and DAL. (b) Detailed illustration of each stage: (1) ECoG recording from subjects, (2) preprocessing including band-pass filtering, notch filtering, and ASR, (3) counterfactual data augmentation through ISBCS between Subject A and Subject B, (4) DAL using EEGNet with task and domain classifiers.}
    \label{fig:SwapAdversarialFramework}
\end{figure}

\subsection{Data augmentation}
\label{sec2.3}
ECoG data exhibit an HDLSS property that often causes unstable convergence and overfitting during adversarial training \citep{sliwowski2023impact}. In computer vision, Mixup \citep{zhang2018mixup} and CutMix \citep{yun2019cutmix} mitigate this by blending cross-domain samples \citep{shankar2018generalizing}. In BCI, representative augmentation methods include channel shuffling \citep{saeed2021learning}, segmentation and recombination \citep{song2023eeg}, and Mixup \citep{rommel2022cadda, mohsenvand2020contrastive}. However, segmentation-and-recombination and Mixup-based methods rely on within-subject transformation and do not reduce inter-subject distributional differences. They also distort temporally meaningful patterns critical for disease diagnosis. Channel shuffling exchanges channels across functionally heterogeneous cortical regions, disrupting spatial-functional correspondence. These limitations are amplified in ECoG, where anatomical differences directly shape electrode-level signal characteristics \citep{haufe2018elucidating, memar2025rise}.

\section{Method}
\label{sec3}

The overall architecture of the proposed SAF is illustrated in Fig.~\ref{fig:SwapAdversarialFramework}. The framework consists of three main stages. First, a preprocessing stage enhances signal fidelity by removing various types of noise from the ECoG recordings. Second, the counterfactual data augmentation stage alleviates structural domain discrepancies caused by inter-subject variability by randomly swapping spatially corresponding channels across subjects. Finally, the DAL stage trains the encoder and domain discriminator adversarially to suppress subject-specific features and learn domain-invariant representations.

\subsection{Preliminary}
\label{sec3.1}

\subsubsection{Problem formulation}
\label{sec3.1.1}
We consider a binary classification problem. In this setting, we use data collected from $L$ individual subjects, denoted as $\{1,2,\ldots,L\}$. For each subject $j$, the dataset is represented as $D_j = \{(x_i^{(j)}, y_i^{(j)}, s_i^{(j)})\}_{i=1}^{n_j}$, where $x_i^{(j)} \in \mathbb{R}^{C \times M}$ denotes the $i$-th ECoG/EEG sample from subject $j$, $y_i^{(j)}$ is its class label e.g., before electrical stimulation (Parkinsonian) or after electrical stimulation (functionally improved condition), and $s_i^{(j)}$ indicates the subject identity. Each sample $x_i^{(j)}$ consists of $C$ channels and $M$ temporal samples, representing a multichannel time series of neural activity. We divide the collected data into source domains $D_S$ and target domains $D_T$. The source domains $D_S = \{D_1, D_2, \ldots, D_K\}$ consist of $K$ domains used for training, while the target domains correspond to unseen domains excluded from training. A feature extractor $f_\theta: \mathbb{R}^{C \times M} \rightarrow \mathbb{R}^d$ and a classifier $g_\phi: \mathbb{R}^d \rightarrow \mathbb{R}^2$ are trained solely on the source domains $D_S$. The objective is for the resulting model $g_\phi \circ f_\theta$ to accurately predict labels $y$ for both the source and unseen target domains.

\subsubsection{Backbone network}
\label{sec3.1.2}
In this study, we adopt EEGNet \citep{lawhern2018eegnet} as the backbone model, which is a lightweight neural network widely used for brain-signal analysis. EEGNet achieves high performance with substantially fewer parameters compared to conventional CNN-based architectures. The model comprises a feature extractor $f_\theta$, which consists of two main blocks, and a classifier $g_\phi$ \citep{cui2023towards}. In the first block of the feature extractor, temporal convolution layers extract frequency-specific features from brain signals, followed by depthwise convolutions that learn spatial representations across channels. The second block applies separable convolutions, which combine depthwise and pointwise operations to model relationships among feature maps. The classifier $g_\phi$, composed of a fully connected layer, predicts the target label $y$ based on the feature representation $z$ obtained from the feature extractor $f_\theta$.

\subsection{Preprocessing}
\label{sec3.2}

\begin{figure}[!t]
    \centering
    \includegraphics[width=0.9\linewidth]{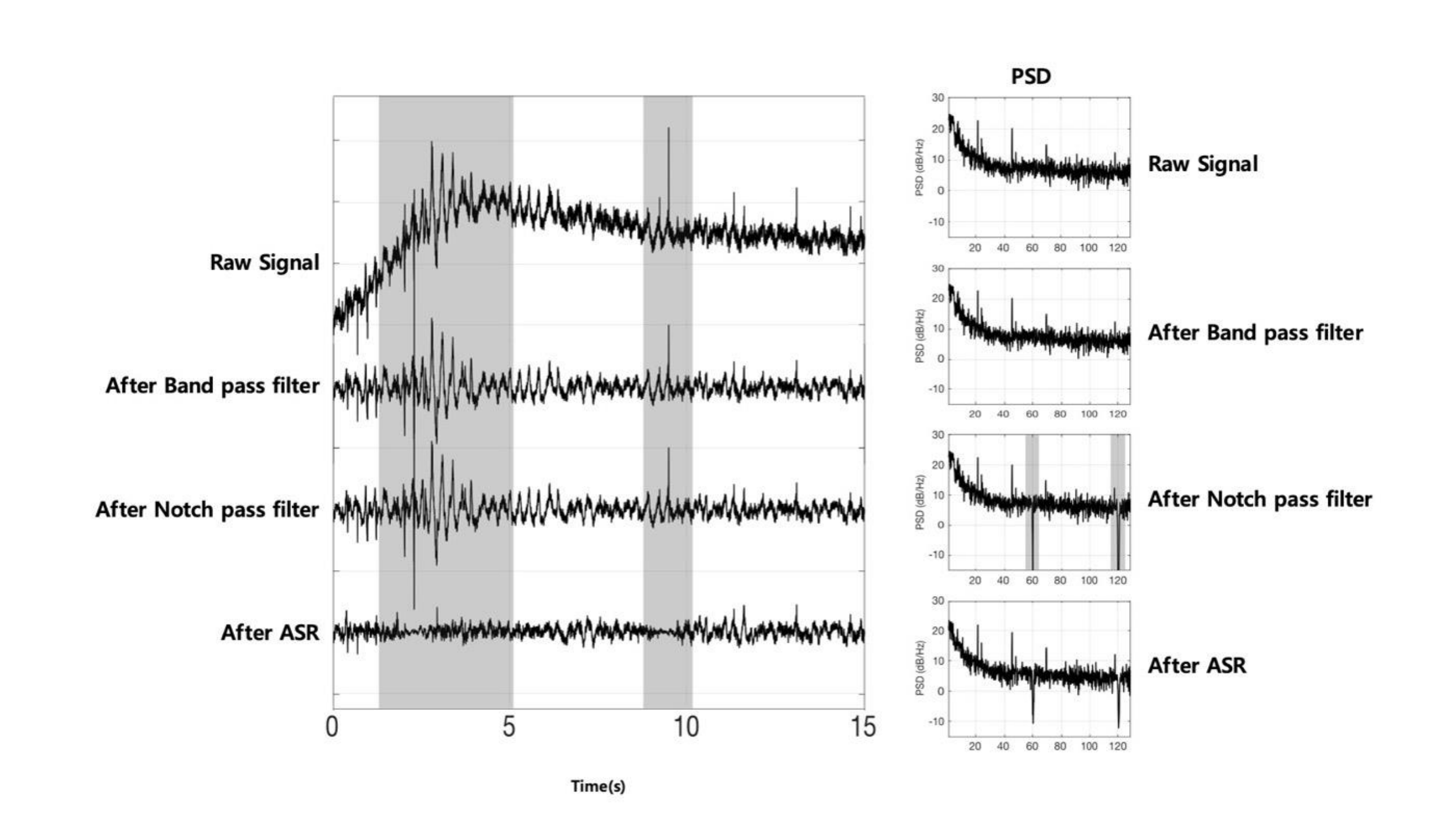}
    \caption{Visualization of the preprocessing pipeline. A representative 15-second segment demonstrating the signal changes during noise removal is presented. Noise removal is applied sequentially using a band-pass filter, a notch filter, and ASR.}
    \label{fig:OverallDataPreprocessing}
\end{figure}

The proposed framework removes artifacts from brain signals through a sequential application of three filtering methods: a band-pass filter, a notch filter, and ASR. The overall process can be expressed as follows:
\begin{equation}
{\overline{x}}=\mathrm{Slicing}(\mathrm{ASR}(\mathrm{Notch}(\mathrm{Bandpass}(x)))).
\end{equation}

All preprocessing steps are applied to the entire length of the ECoG recordings at once, after which the processed data are segmented according to a predefined window size. Descriptions of each method are provided below.

\textbf{Band-pass filtering.} A band-pass filter between 1 Hz and 128 Hz was applied to remove low-frequency noise associated with eye blinks and muscle movements, as well as high-frequency noise and detrending effects.

\textbf{Notch filtering.} To suppress power-line interference, notch filters at 60 Hz and its harmonic (i.e., 120 Hz) were applied, selectively removing those specific frequency components.

\textbf{ASR.} To eliminate unpredictable and non-repetitive artifacts, ASR was applied. It detects clean calibration segments with low noise variance to estimate a reference subspace and removes components that exceed a root mean square (RMS)-based threshold, reconstructing the signal from the remaining components \citep{chang2018evaluation}.  

As shown in Fig.~\ref{fig:OverallDataPreprocessing}, we visualize the denoising process of a representative sample step by step in both the time and frequency domains. The band-pass filter removed low-frequency drifts and high-frequency noise, while the notch filter suppressed the 60 Hz and 120 Hz interference. Finally, ASR eliminated artifacts between 1--5 s and around 10 s.

\subsection{Domain generalization}
\label{sec3.3}

\begin{figure*}[t]
    \centering
    \begin{minipage}[t]{0.67\textwidth}
        \centering
        \includegraphics[width=\linewidth]{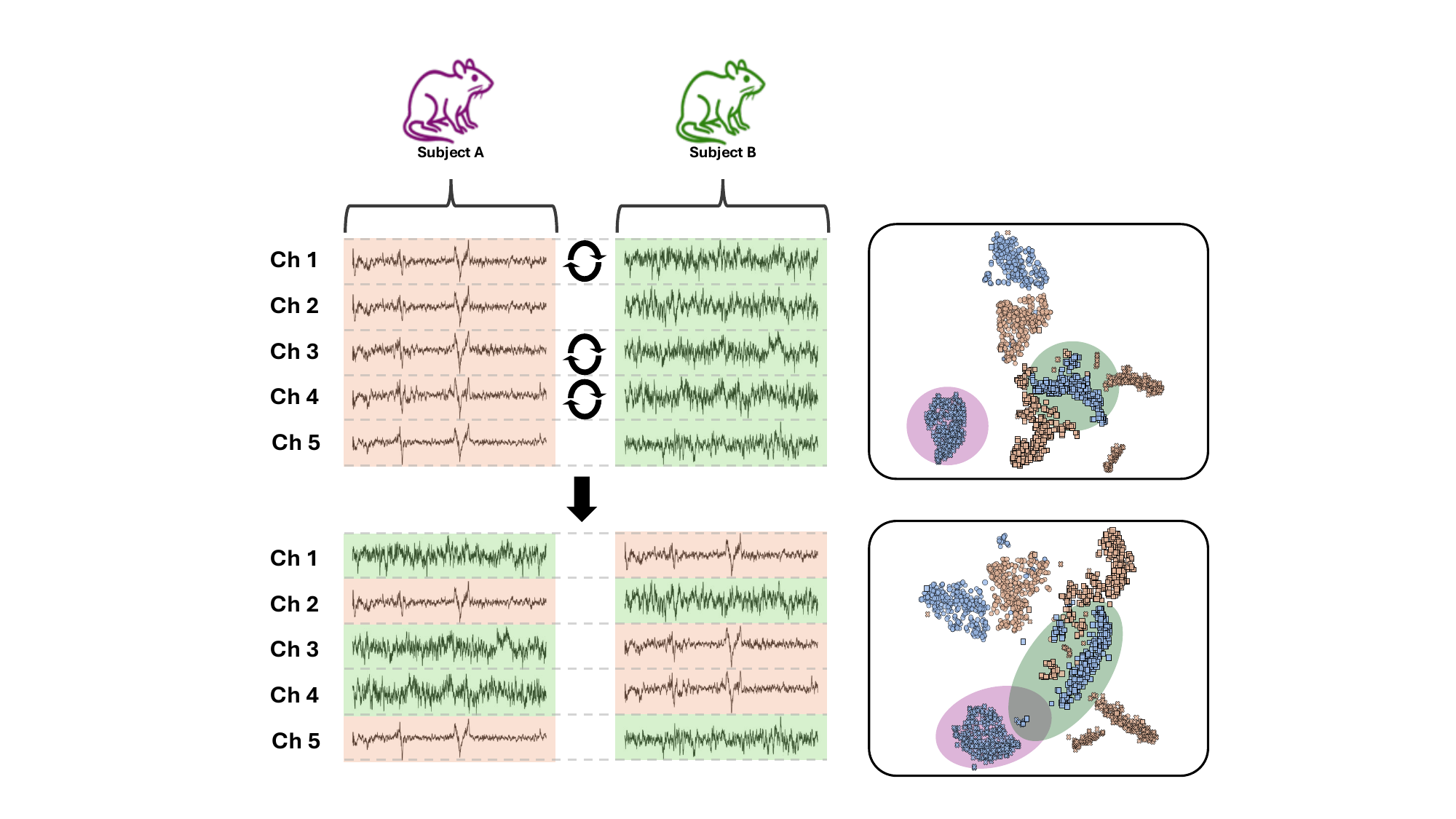}
        \caption{ISBCS: A counterfactual data augmentation method that randomly swaps channel data between subjects A and B within the same class. The t-SNE visualization shows that ISBCS induces overlapping distributions between subjects, reducing the inter-subject distributional separability. Here, the red and green circles represent different subjects, respectively.}
        \label{fig:ISBCS}
    \end{minipage}%
    \hspace{0.03\textwidth}%
    \begin{minipage}[t]{0.3\textwidth}
        \centering
        \includegraphics[width=\linewidth]{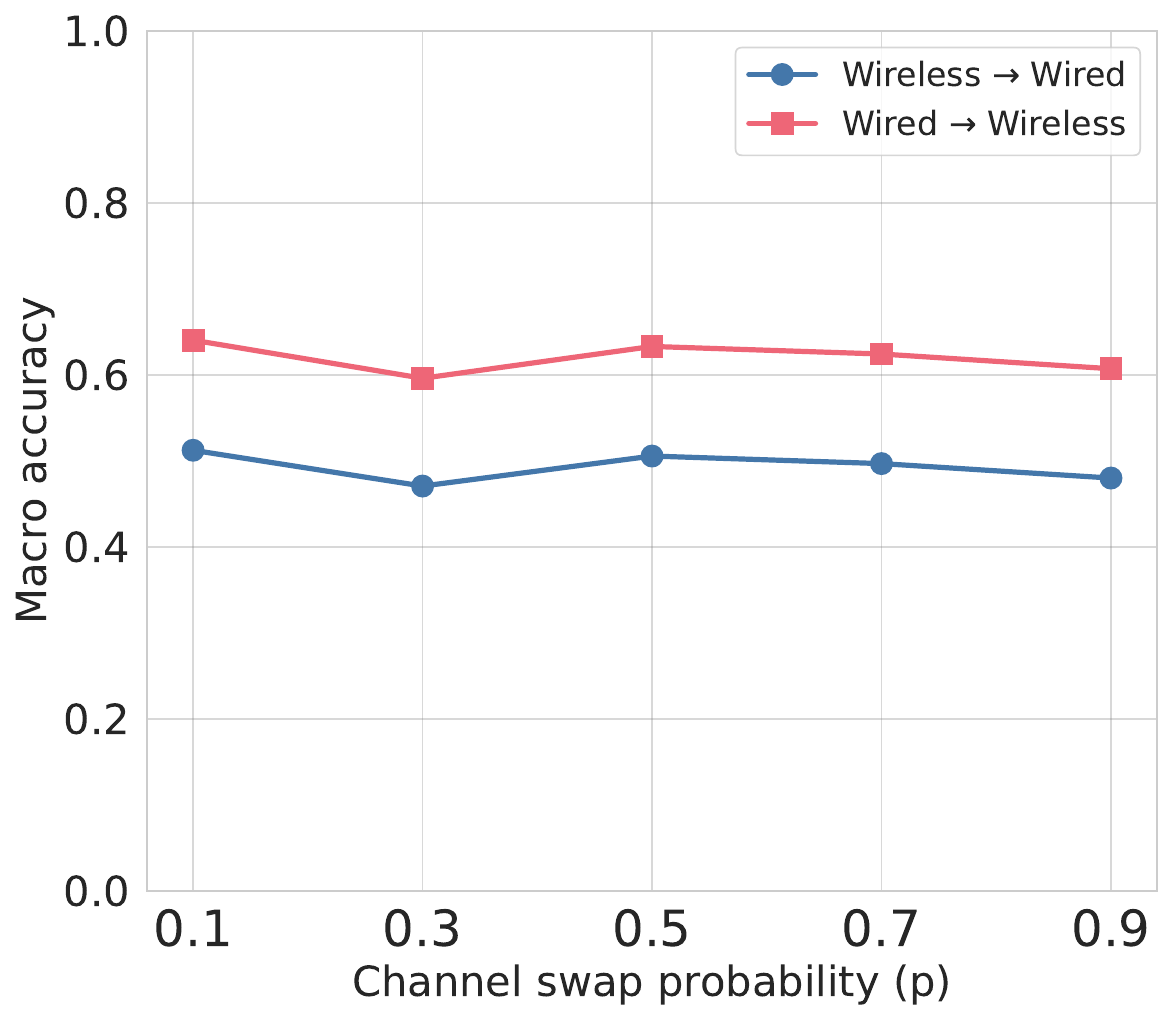}
        \caption{Sensitivity analysis of the ISBCS channel-swap probability $p$. Blue denotes the results of training on wireless and testing on wired data, whereas red denotes the results of training on wired and testing on wireless data.}
        \label{fig:ISBCSsensitivity}
    \end{minipage}
\end{figure*}

\subsubsection{Counterfactual data augmentation}
\label{sec3.3.1}

ECoG signals suffer from large inter-subject variability caused by individual biological differences, which hinders model generalization \citep{huang2023discrepancy}. To mitigate this structural bias and enhance model generalization, we propose a novel data augmentation method called ISBCS.
This method is a bio-inspired data augmentation method in which swapping is performed only between spatially corresponding channels across subjects. {
Here, spatially corresponding channels refer to channels recorded from graphene-based electrodes placed at the same predefined anatomical coordinates across rats. Based on the motor-cortex brain map, electrodes at the same predefined anatomical coordinates in each rat were considered to be functionally corresponding \citep{shin2025wireless, leergaard2000somatotopic, penfield1937somatic}.
}

Let $x_i \in \mathbb{R}^{C \times M}$ denote the preprocessed original sample used as the anchor for ISBCS, with subject identity $s_i$ and class label $y_i$, where $C$ is the number of channels and $M$ is the number of temporal samples. ISBCS operates through the following three steps. First, in the \textit{channel selection}, for each channel $c \in \{1, \ldots, C\}$, a binary indicator $b_c$ is independently sampled from a Bernoulli distribution with probability $p$ (i.e., $b_c \sim \text{Bernoulli}(p)$). Only channels with $b_c = 1$ are selected for swapping. This process generates samples composed of channels from two different subjects, which can be interpreted as forming an intermediate distribution between the two subjects and thereby helping reduce the distribution gap across subjects. Second, in the \textit{subject pairing}, for each selected channel, a random permutation $\pi_c$ of the subject indices $\{1, \ldots, K\}$ is generated independently, defining subject pairs $(s, \pi_c(s))$ for channel data exchange. The paired subjects are selected randomly. This allows the method to form intermediate distributions from diverse subject pairs, rather than being biased toward any specific pair. Finally, in the \textit{class-consistent swap}, within each subject pair, swapping is performed only within the same class so that class-discriminative information is preserved. This prevents the generation of augmented samples containing mixed-class signals. The augmented dataset $D_S^\prime$ has the same size as the original dataset $D_S$ (i.e., $|D_S^\prime| = N$, where $N = \sum_{k=1}^{K} n_k$ is the total number of source samples). ISBCS replaces each original sample in place with its augmented counterpart, rather than generating additional samples or combining the two. The resulting ISBCS-augmented sample $x_i^\prime$ is formally defined channel-wise as:

\begin{equation}
x_{i,c}^\prime =
\begin{cases}
x_{j,c} & \text{if } b_c = 1,\ s_j = \pi_c(s_i),\ \text{and } y_i = y_j \\
x_{i,c} & \text{otherwise}
\end{cases},
\end{equation}
where $j$ denotes the sample index paired with $i$ under the subject permutation $\pi_c$ (i.e., $s_j = \pi_c(s_i)$), and the class-consistency constraint $y_i = y_j$ ensures that swapping occurs only between same-class samples, guaranteeing that label semantics are preserved.
\aqrevision{Figure~\ref{fig:ISBCS} (left) illustrates a five-channel example in which
selected spatially corresponding channels are randomly swapped between
same-class samples from subjects A and B.
Figure~\ref{fig:ISBCS} (right) visualizes the changes in the feature
distributions for each subject.}
The only hyperparameter \aqrevision{of ISBCS} is the channel-swap probability $p$,
which controls the expected proportion of channels involved in swapping.
\aqrevision{Figure~\ref{fig:ISBCSsensitivity} shows the sensitivity of macro accuracy
to the channel-swap probability $p$ in the cross-cohort evaluations.
The results show that macro accuracy remained relatively stable across
the evaluated values of $p$.}

For domain-adversarial learning, the subject label of each ISBCS-augmented sample is defined as $s_i$, corresponding to the subject identity of the anchor sample. Because the subject permutation is independently determined for each selected channel, a single ISBCS-augmented sample may contain channels originating from multiple paired subjects. The subject label of ISBCS-augmented sample is retained as $s_i$, corresponding to the subject identity of the original sample before channel swapping. For example, if Subject A is used as the anchor, different selected channels may be exchanged with corresponding channels from Subjects B, C, or D, while the resulting ISBCS-augmented sample retains Subject A as its subject label.

{
For DAL, domain labels can be defined using two strategies (i.e., hard domain labeling and soft domain labeling). Hard domain labels retain the anchor sample's domain label, whereas soft domain labels are defined based on the proportion of the total channels in the ISBCS-augmented sample that originated from each subject. Based on a sensitivity analysis comparing the two strategies, we adopted hard domain labels, which yielded higher performance on average than soft domain labels under the evaluated conditions. The detailed results of this sensitivity analysis are provided in Supplementary Table S1.
}

Finally, the augmented data are used to train the classification model as follows:

\begin{equation}
L_{task}=E_{(\overline{x},y)\sim D_S^\prime}L_{CE}(g_\phi(f_\theta(\overline{x})),y),
\end{equation}
where $D_S^\prime={({\overline{x}}_i,y_i,s_i)}_{i=1}^N$ denotes the augmented dataset generated by the ISBCS method, and $L_{CE}$ is the cross-entropy loss.

\subsubsection{Domain-adversarial learning}
\label{sec3.3.2}

ISBCS mitigates subject-specific bias at the data level by swapping channels between samples from different subjects of the same class. We further introduce DAL to suppress subject-dependent features, yielding domain-invariant representations and improved generalization to unseen domains.

To this end, we design the architecture for DAL inspired by the approach of \citet{kim2019learning}. The framework combines two complementary objectives: (1) a domain-adversarial loss $L_{domain}$ that trains the feature extractor to confuse a domain classifier, and (2) a mutual information (MI) suppression loss $L_{MI}$ that explicitly penalizes subject-discriminative structure in the feature space. Specifically, a domain classifier $h_\psi$ is employed to predict the subject identity $s$ from the feature representation $z = f_\theta(\overline{x})$. The augmented data $D_S^\prime$ are fed into the feature extractor, and the resulting features are passed through a gradient reversal layer (GRL) \citep{ganin2016domain}, denoted by $\mathrm{GRL}(\cdot)$, before being fed to the domain classifier. The GRL performs an identity mapping during the forward pass, as defined in Eq.~(\ref*{eq:grl_forward}), and multiplies the gradient of $L_{domain}$ propagated to the feature extractor by $-\lambda_{GRL}$ during backpropagation, as defined in Eq.~(\ref*{eq:grl_backward}), enabling adversarial learning.

{
\begin{subequations}\label{eq:grl_operations}
\begin{align}
&\mathrm{GRL}\bigl(f_\theta(\overline{x})\bigr) = f_\theta(\overline{x})
&& \text{(forward pass)}, \label{eq:grl_forward} \\
&\frac{\partial L_{domain}}{\partial f_\theta(\overline{x})} = -\lambda_{GRL}\frac{\partial L_{domain}}{\partial \mathrm{GRL}\bigl(f_\theta(\overline{x})\bigr)}
&& \text{(backward pass)}. \label{eq:grl_backward}
\end{align}
\end{subequations}
}
As a result, the domain classifier learns to predict subject labels accurately, while the feature extractor learns to suppress subject-specific information. The corresponding objective is formulated as:

{
\begin{equation}\label{eq:domain_loss}
L_{domain} = \min_{\psi} \max_{\theta} \mathbb{E}_{(\overline{x}, s) \sim D_S^\prime} L_{CE}\bigl(h_\psi\bigl(\mathrm{GRL}\bigl(f_\theta(\overline{x})\bigr)\bigr), s\bigr).
\end{equation}
}
In addition, we introduce $L_{MI}$ as a surrogate objective to mitigate the MI between $z$ and $s$. According to \citet{kim2019learning}, the MI can be decomposed as:

\begin{equation}
I(Z; S) = H(S) - H(S \mid Z),
\end{equation}
where $H(S)$ is the marginal entropy of the subject labels and $H(S \mid Z)$ is the conditional entropy of $S$ given $Z$. Since $H(S)$ is determined solely by the training set and remains constant during optimization, minimizing $I(Z; S)$ is equivalent to maximizing $H(S \mid Z)$, which can be expressed as:

\begin{equation}
H(S \mid Z) = \mathbb{E}_{z} \left[ -\sum_{k=1}^{K} p(s_k \mid z) \log p(s_k \mid z) \right],
\end{equation}
where $K$ is the number of subjects. Since $h_\psi$ is trained to approximate the posterior $p(S \mid z)$, we define the MI suppression loss as:

\begin{equation}
L_{MI} = \mathbb{E}_{z \sim f_\theta(\overline{x})} H(h_\psi(z)), \quad \text{where} \quad H(q) = -\sum_{k=1}^{K} q_k \log q_k.
\end{equation}

Maximizing $L_{MI}$ encourages $h_\psi(z)$ to approach a uniform distribution over subjects, implying that $z$ carries less discriminative information about subject identity---i.e., higher conditional entropy and consequently lower MI. Note that this loss serves as a surrogate objective based on conditional entropy maximization rather than a rigorous MI estimator; it complements the GRL-based adversarial loss by preventing overconfident domain predictions.
Finally, the overall objective is defined as follows:

{
\begin{equation}\label{eq:total_loss}
L_{total} = L_{task} + L_{domain} - \lambda_{MI}L_{MI}.
\end{equation}
}

Here, $\lambda_{MI}$ controls the contribution of $L_{MI}$, whereas $\lambda_{GRL}$ controls the magnitude of the reversed gradient propagated from the domain classifier to the feature extractor through the GRL during backpropagation. Together, these two mechanisms help suppress subject-specific information, thereby encouraging the feature extractor to learn domain-invariant representations that generalize well to unseen domains. To optimize these objectives, training is performed in two steps with distinct optimization directions for each component. First, with the domain classifier $h_\psi$ fixed, the feature extractor $f_\theta$ is optimized to minimize $L_{task} - \lambda_{MI}L_{MI}$, while the task classifier $g_\phi$ is optimized to minimize $L_{task}$. The gradient of $L_{MI}$ is propagated only to $f_\theta$. Second, the domain classifier $h_\psi$ is optimized to minimize $L_{domain}$, whereas the feature extractor $f_\theta$ is adversarially optimized through the GRL to maximize $L_{domain}$. The task classifier $g_\phi$ is not updated in this step.

\section{Datasets}
\label{sec4}

\begin{figure*}[!t]
    \centering
    \includegraphics[width=0.95\textwidth]{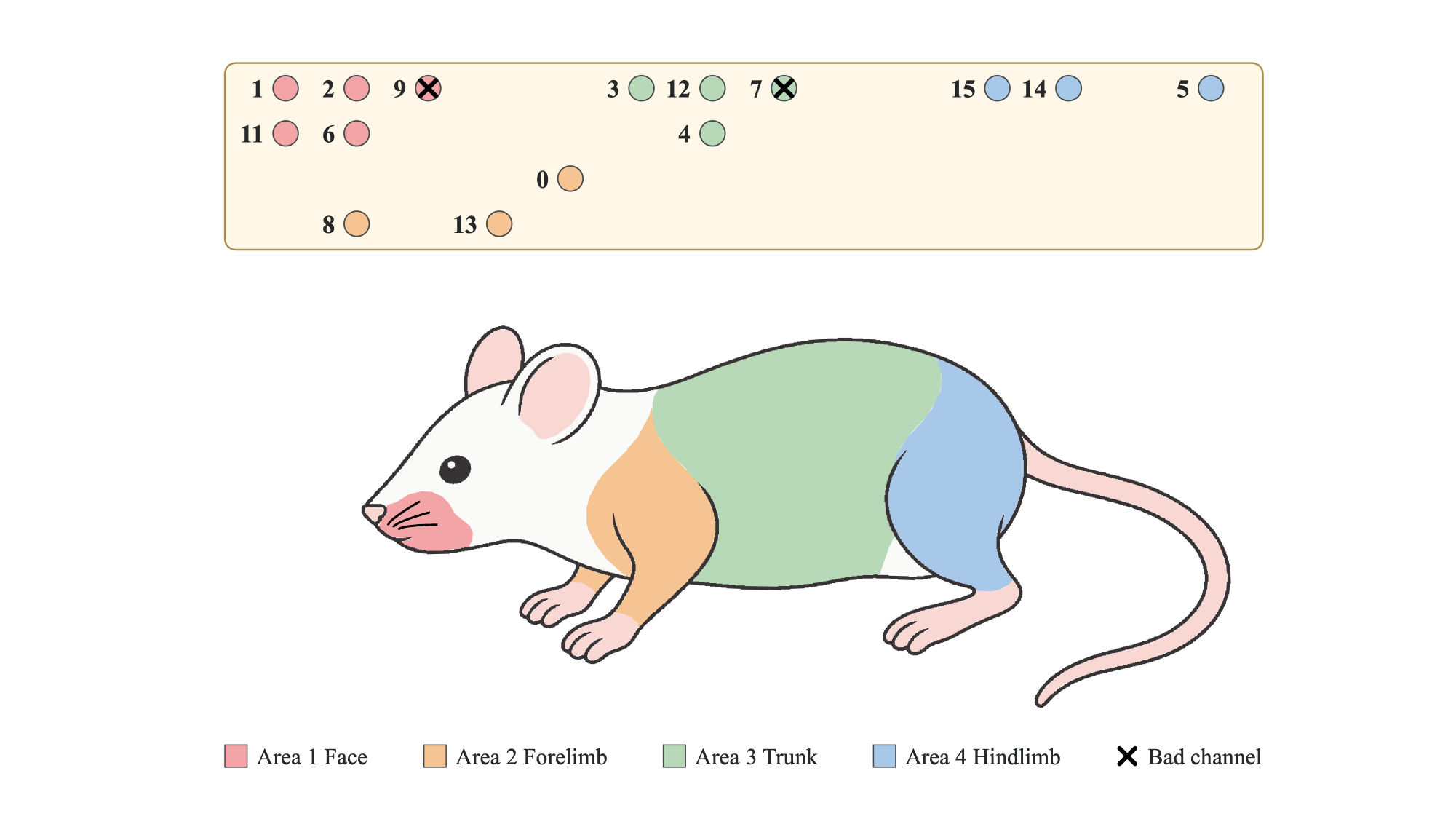}
    \caption{ECoG channel layout and corresponding body regions. Matching colors indicate the correspondence between channel groups and body regions based on the motor-cortex brain map. Black crosses indicate bad channels (7 and 9).}
    \label{fig:ECoGChannelLayout}
\end{figure*}

\subsection{ECoG dataset}
\label{sec4.1}

\begin{figure*}[t]
    \centering
    \begin{tabular}{@{}p{0.52\textwidth}@{\hspace{0.02\textwidth}}p{0.46\textwidth}@{}}
        \begin{minipage}[t]{\linewidth}
            \vspace{0pt}
            \centering
            \includegraphics[width=\linewidth]{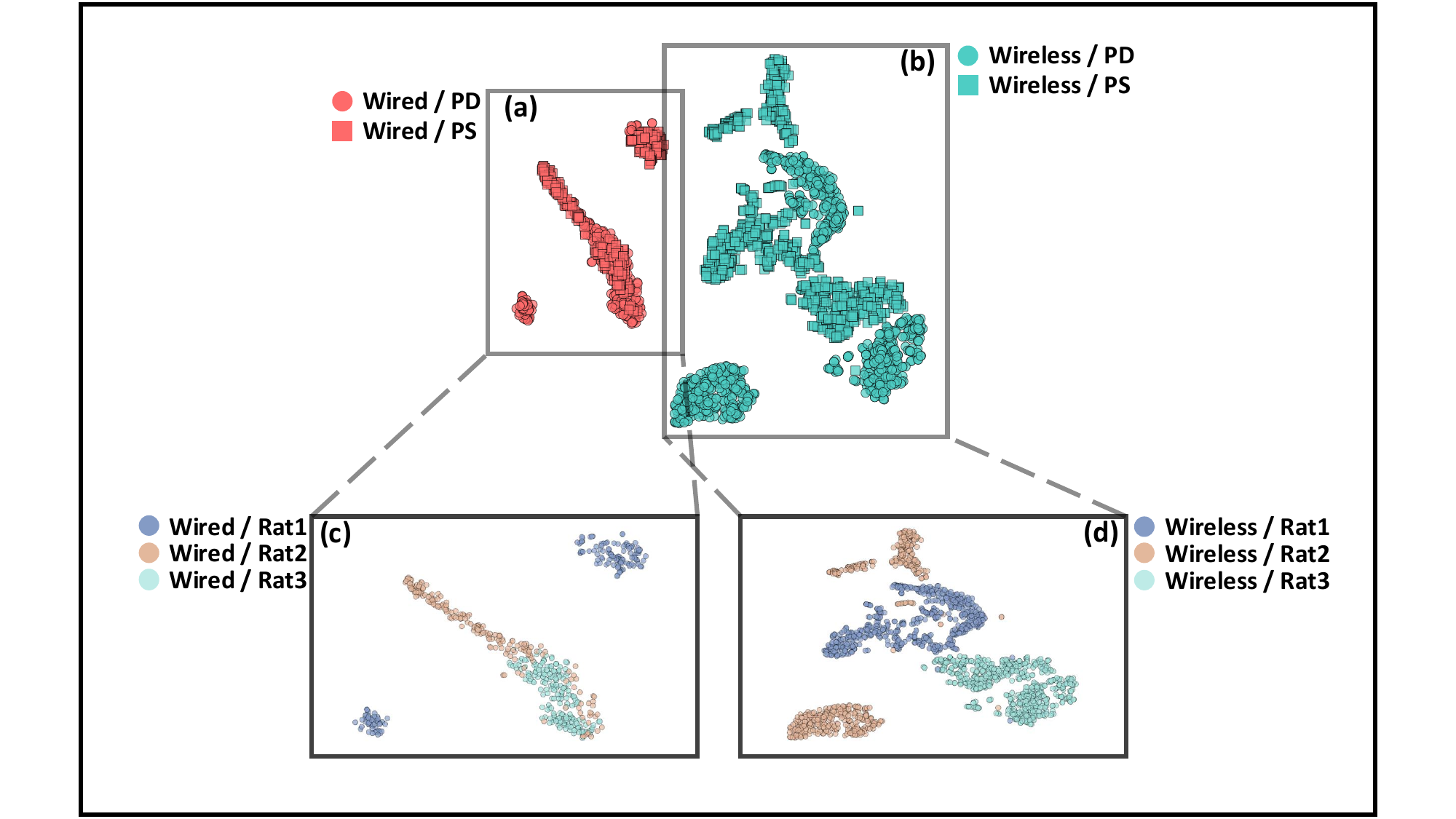}
            \caption{T-SNE visualization of the ECoG data after PSD transformation. (a), (b) Distributions of all subjects in the wired and wireless datasets, respectively. (c) Distributions of individual subjects in the wired dataset. (d) Distributions of individual subjects in the wireless dataset. PS denotes the state after electrical stimulation, while PD denotes Parkinson's disease.}
            \label{fig:ECoGPsdTsne}
        \end{minipage}
        &
        \begin{minipage}[t]{\linewidth}
            \vspace{0pt}
            \centering
            \includegraphics[width=\linewidth]{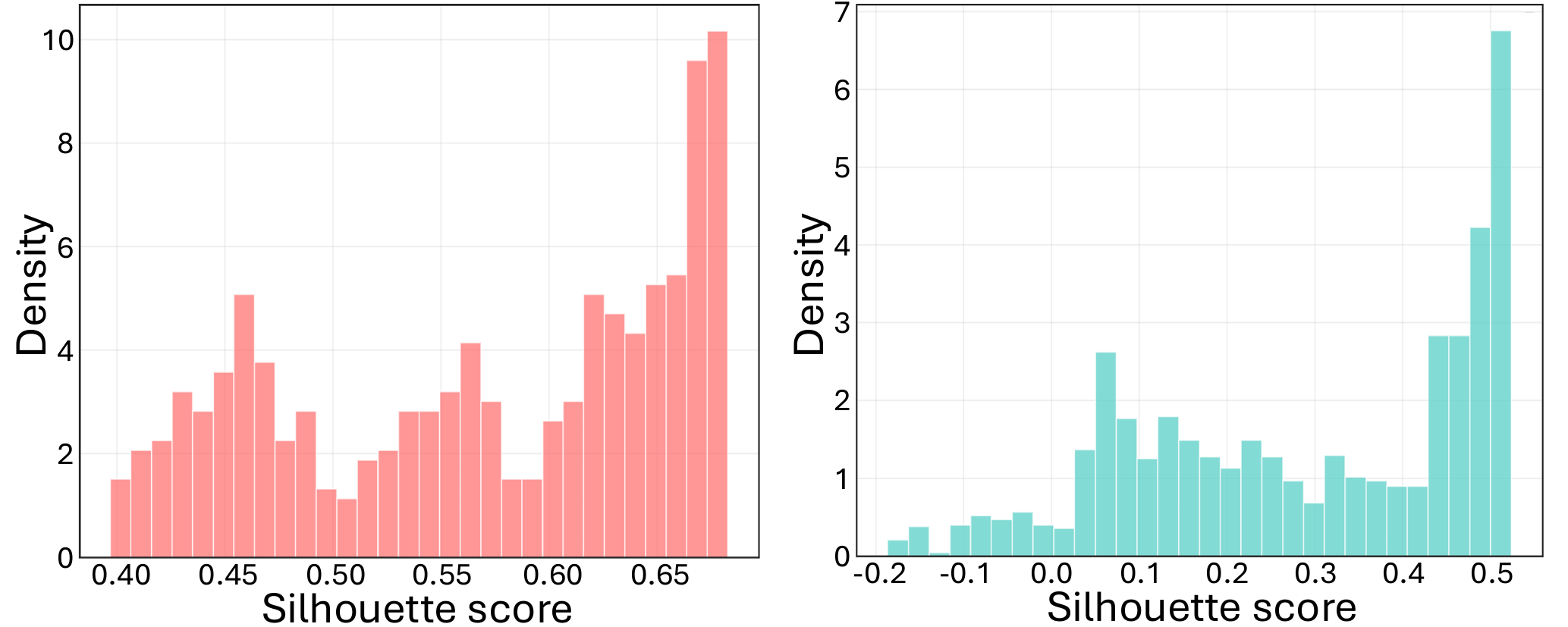}
            \caption{Silhouette score histograms based on t-SNE visualization of the ECoG data. (Left) Wireless ECoG dataset. (Right) Wired ECoG dataset.}
            \label{fig:ECoGSSHistogram}

            \vspace{0.8em}
            {\captionsetup{type=table}
            \caption{Number of samples in the ECoG dataset.}
            \label{table:ECoGSample}}
            \sctablesetup
            \setlength{\tabcolsep}{2pt}
            \resizebox{\linewidth}{!}{%
            \begin{tabular}{lllllll}
                \toprule
                 & \multicolumn{3}{l}{Wireless} & \multicolumn{3}{l}{Wired} \\
                \cmidrule(lr){2-4} \cmidrule(lr){5-7}
                 & Rat 1 & Rat 2 & Rat 3 & Rat 4 & Rat 5 & Rat 6 \\
                \midrule
                \makecell[l]{Class 0\\(Before electrical stimulation)} & 308 & 305 & 301 & 78 & 99 & 69 \\
                \makecell[l]{Class 1\\(After electrical stimulation)} & 300 & 242 & 331 & 114 & 99 & 99 \\
                \bottomrule
            \end{tabular}
            }
        \end{minipage}
    \end{tabular}
\end{figure*}

\subsubsection{Data acquisition from previous study}
\label{sec4.1.1}

To construct the training and evaluation datasets for the ECoG-based classification model, we utilized the data collected in a previous study \citep{shin2025wireless}. Specifically, PD was induced in eleven healthy rats using the 6-OHDA method, and ECoG signals were recorded from seven of them. One subject with severe noise contamination was excluded, and the remaining six rats were used for the experiments.

{
Among the six rats, ECoG recordings were obtained from three rats using a wired system for 12 days and from the other three rats using a wireless system for 7 days. In both recording setups, the rats were freely moving inside a $43 \times 33 \times 30$~cm cage. The sampling rates for the wired and wireless recordings were 30,000~Hz and 512~Hz, respectively.
}

{
During data acquisition in the previous study\citep{shin2025wireless}, electrode implantation was performed using standardized stereotaxic coordinates and the same electrode configuration across all animals. The skull was prepared using stereotaxic reference points, and the electrode array was positioned over the motor cortex according to the predefined anatomical coordinates. A standardized electrode implantation procedure was used to minimize variability in electrode placement across rats. In addition, motor threshold testing was performed after implantation, and stimulation-induced motor responses were confirmed. These responses provided functional confirmation that the electrode array was targeting the motor cortical region.
}

\subsubsection{Ethics statement} Adult male Sprague--Dawley rats weighing 260--300~g were used. The rats were housed in a room maintained at 25~$^\circ$C under a 12-h light/12-h dark cycle. The rats were obtained from DBL (Daehan Bio Laboratory, Eumseong, Republic of Korea). All animal procedures were approved by the Institutional Animal Care and Use Committee at Incheon National University (INU-ANIM-2017-08) in accordance with their guidelines. At the end of all experiments, the rats were euthanized by inhalation of an overdose of 5\% isoflurane using an induction chamber.

\begin{figure*}[!t]
    \centering
    \includegraphics[width=0.9\textwidth]{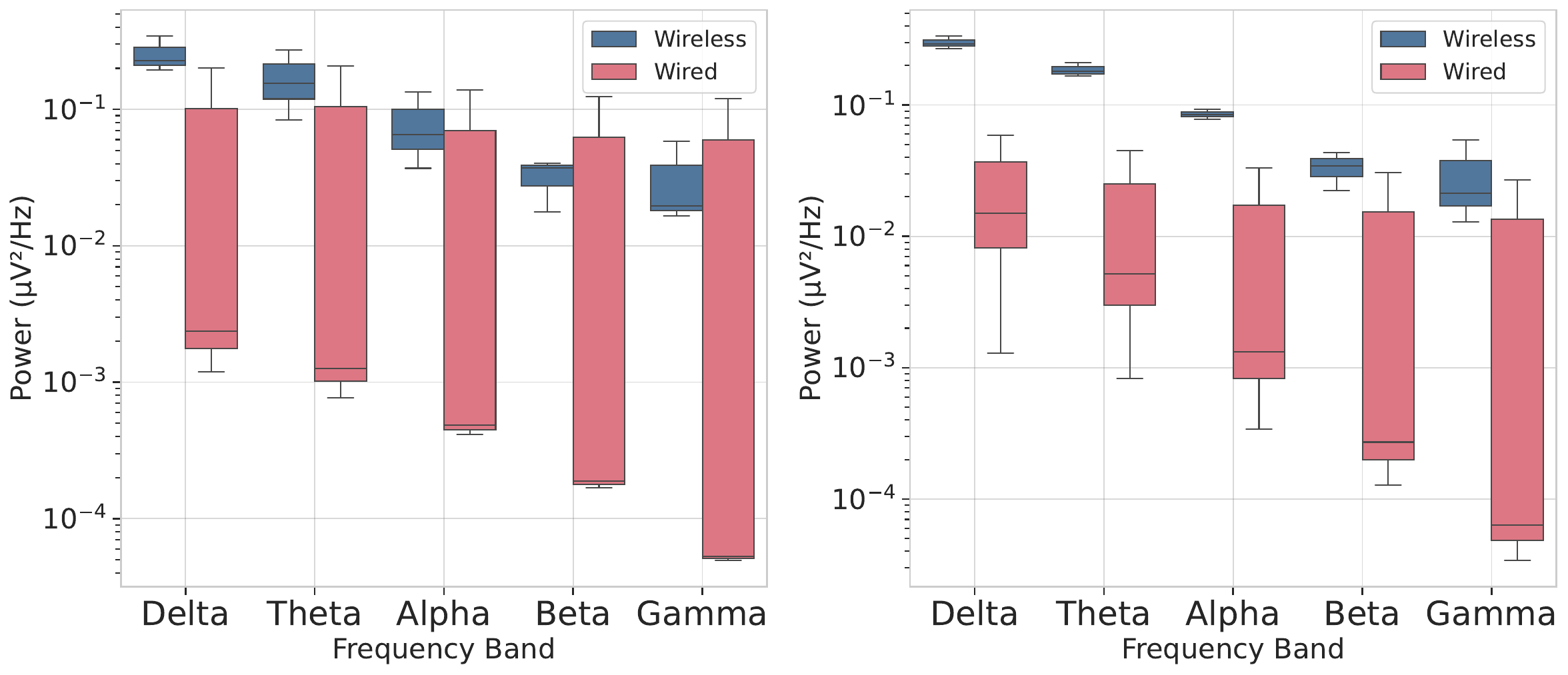}
    \caption{Power spectral density analysis across different datasets and classes: Comparison of frequency band-specific PSD between wireless and wired in {RATCOP} dataset. (Left) Before electrical stimulation. (Right) After electrical stimulation. Blue and red denote the wireless and wired dataset, respectively. Note that the y-axis is shown on a logarithmic scale for better visualization.}
    \label{fig:PSDAnalysisMOCOP}
    
    \vspace{1em}
    \includegraphics[width=0.9\textwidth]{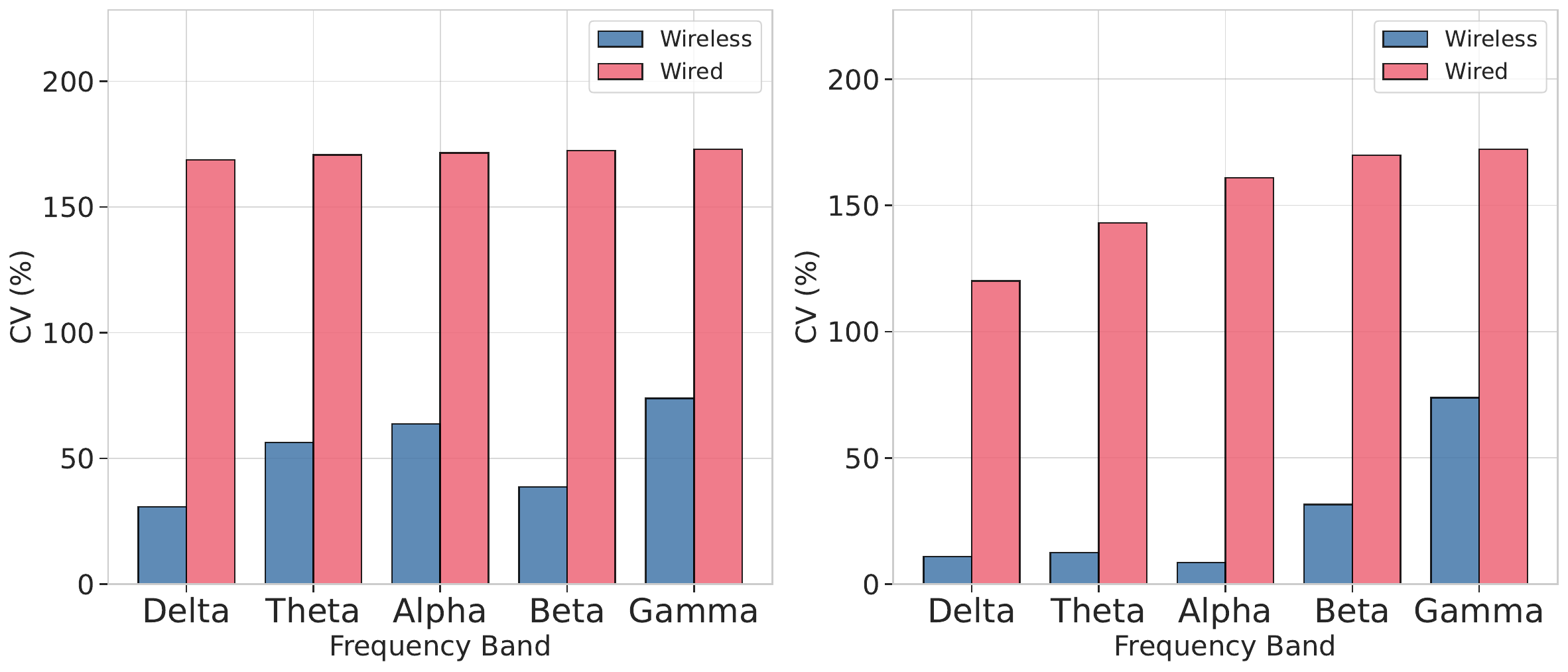}
    \caption{Inter-subject variability analysis using coefficient of variation (CV): Comparison of inter-subject variability between wireless and wired in {RATCOP} dataset. (Left) Before electrical stimulation. (Right) After electrical stimulation. Blue and red denote the wireless and wired dataset, respectively.}
    \label{fig:PSDCVAnalysisMOCOP}
\end{figure*}

\subsubsection{Dataset labeling and construction}
\label{sec4.1.2}

For the same subjects, electrical stimulation was applied to the motor cortex using a graphene-based electrode, as illustrated in Fig.~\ref{fig:SwapAdversarialFramework} (b).
Electrical stimulation was administered once per day to all rats. After one week of stimulation, the rats that received stimulation showed notable improvement in motor function, as confirmed by gait and other behavioral tests, compared to the unstimulated state. In addition, the ECoG recordings showed a decrease in high-frequency activity and an increase in low-frequency components after electrical stimulation. Accordingly, the dataset includes ECoG signals corresponding to neural states before and after electrical stimulation in the same 6-OHDA-induced PD rat models. The detailed procedure for ECoG data acquisition is provided in the previous study \citep{shin2025wireless}.

A total of six rats were used in the experiments. For each subject, the state without electrical stimulation was defined as class 0, and the state after one week of continuous stimulation was defined as class 1, formulating the task as a binary classification problem. The class labels were determined based on behavioral tests conducted before and after electrical stimulation, following the procedure described in our previous study \citep{shin2025wireless}.

{
ECoG recordings were originally acquired from 15 channels (graphene-based electrodes) in both recording setups. Inspection of the raw signals identified channels 7 and 9 as bad channels because they exhibited persistent noise in some recordings compared with the other channels. To maintain the same channel configuration across rats, these two channels were excluded from the recordings of all rats. The remaining 13 channels were used in all subsequent analyses and are included in the released dataset.
}

{
Figure~\ref{fig:ECoGChannelLayout} shows the channel layout of the ECoG dataset and the body regions corresponding to each channel based on the motor-cortex brain map \citep{shin2025wireless, leergaard2000somatotopic, penfield1937somatic}. The upper part of Figure~\ref{fig:ECoGChannelLayout} shows the layout of the graphene-based electrodes used for ECoG data acquisition and the channel index corresponding to each electrode. Black crosses mark bad channels 7 and 9 to distinguish them from the remaining channels used in the experiments. Based on the rat brain map, the lower part shows the body regions corresponding to the functional regions of the brain at each electrode coordinate. Corresponding channels and body regions are shown in matching colors, and the legend explains the meaning of each color. For example, channels 5, 14, and 15 are shown in blue, as is the corresponding body region (Hindlimb).
}

The wired ECoG data were downsampled from 30,000~Hz to 512~Hz, and both wired and wireless recordings were divided into 6-second non-overlapping segments. {Table~\ref{table:ECoGSample} presents the number of samples for each class and subject in the proposed ECoG dataset.} The samples from each subject were split into training, validation, and test sets at a ratio of 8:1:1. These predefined sets were used according to the experimental setting, as described in the Experiments section. Across all settings, only the training and validation sets of the source data were used for model training and selection, whereas the predefined test sets of the target data were used exclusively for final evaluation.

\subsubsection{Dataset distribution analysis}
\label{sec4.1.3}

The proposed ECoG dataset was collected over an extended period while the rats were freely moving, which inevitably introduced movement artifacts and power-line noise. In addition, biological differences among individual rats lead to subject-specific variability. To analyze the biases across subjects and acquisition modalities, we computed the power spectral density (PSD) \citep{kyllonen2024unsupervised} of each recording and visualized the feature representations using t-SNE. 

Figure~\ref*{fig:ECoGPsdTsne} (a) and (b) show that the wired and wireless cohorts form clearly separated clusters, indicating a substantial compound distributional difference between the two acquisition modalities. Because the datasets were obtained from different rats and differed in acquisition system, recording duration, and original sampling rate, this separation cannot be attributed to acquisition system alone. Figure~\ref{fig:ECoGPsdTsne} (c) and (d) show distinct inter-subject clustering within each modality. In addition, as shown in Fig.~\ref{fig:ECoGSSHistogram}, we quantitatively evaluated the degree of cluster separation using the silhouette score. The results show clear inter-subject clustering in both wired and wireless datasets, with greater subject-level variability observed in the wireless data.

Moreover, we quantitatively analyzed the PSD distributions across frequency bands and the inter-subject variability within the wired and wireless datasets. As shown in Fig.~\ref{fig:PSDAnalysisMOCOP}, both datasets exhibited higher PSD values in the low-frequency bands (i.e., Delta, Theta, Alpha) than in the high-frequency bands (i.e., Beta, Gamma). In both modalities, the PSD distributions differed before and after electrical stimulation, indicating meaningful class-wise differences. Notably, the wireless data showed higher PSD values in the low-frequency bands, whereas the wired data exhibited relatively lower power. After electrical stimulation, the wireless data showed an increase in $\delta$-band power, while the wired data showed a decrease in power across all frequency bands. These results highlight a clear compound domain shift between the wired and wireless cohorts and motivate evaluation of generalization across cohort and acquisition-related differences.

Figure~\ref{fig:PSDCVAnalysisMOCOP} reports the coefficients of variation (CV) of the mean PSD across subjects. In the PD condition, both wired and wireless datasets exhibited high variability across subjects. After electrical stimulation, variability in the wireless data decreased slightly in the low-frequency bands but remained substantial, whereas the wired data maintained consistently high variability across all bands. These results indicate substantial inter-subject variability and the need for generalization across subjects within each modality.

\begin{figure*}[!t]
    \centering
    \begin{tabular}{@{}p{0.52\textwidth}@{\hspace{0.02\textwidth}}p{0.46\textwidth}@{}}
        \begin{minipage}[t]{\linewidth}
            \vspace{0pt}
            \centering
            \includegraphics[width=\linewidth]{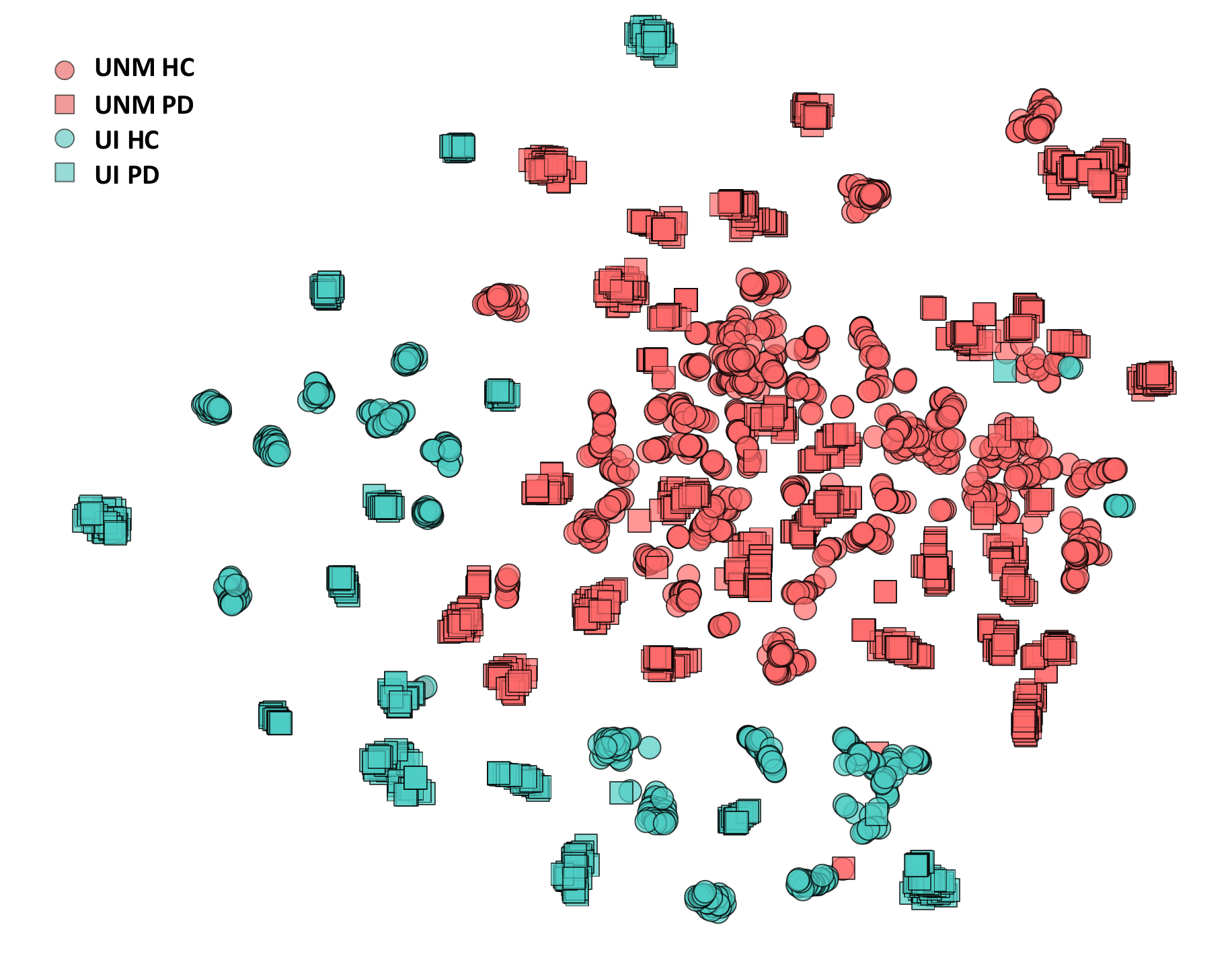}
            \caption{T-SNE visualization of the UI and UNM dataset after PSD transformation.}
            \label{fig:EEGPsdTsne}
        \end{minipage}
        &
        \begin{minipage}[t]{\linewidth}
            \vspace{0pt}
            \centering
            \includegraphics[width=\linewidth]{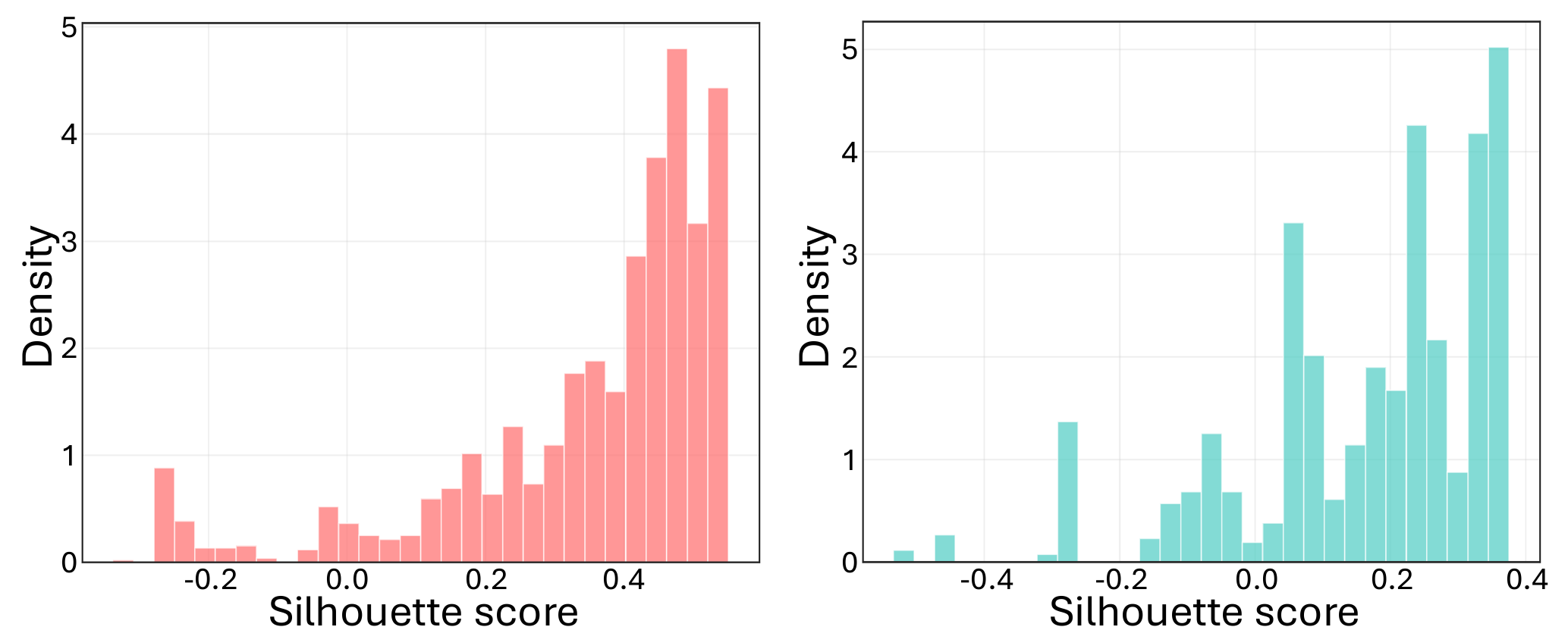}
            \caption{Silhouette score histograms based on t-SNE visualization of the EEG data. (Left) UNM dataset. (Right) UI dataset.}
            \label{fig:EEGSSHistogram}

            \vspace{0.8em}
            {\captionsetup{type=table}
            \caption{Number of samples in the EEG datasets.}
            \label{table:EEGSample}}
            \sctablesetup
            \begin{tabular}{lll}
                \toprule
                 & UI & UNM \\
                \midrule
                Number of subject & 28 & 52 \\
                Class 0 (Healthy control) & 429 & 921 \\
                Class 1 (Parkinson's disease) & 438 & 838 \\
                \bottomrule
            \end{tabular}
        \end{minipage}
    \end{tabular}
\end{figure*}

\subsection{EEG dataset}
\label{sec4.2}

\subsubsection{Data acquisition from previous study}
\label{sec4.2.1}

We utilized two publicly available EEG datasets that are widely used for PD classification \citep{cavanagh2018diminished, singh2020frontal, sugden2023generalizable}: the University of New Mexico (UNM) and University of Iowa (UI) datasets. Both are commonly adopted as benchmark datasets for EEG-based Parkinson's disease diagnosis. A detailed description of each dataset is as follows:

\begin{itemize}
    \item \textbf{UNM Dataset}: This dataset consists of 52 participants, including 27 patients with PD and 25 healthy controls. EEG signals were collected both while the patients were on dopaminergic medication and after a 12-hour medication withdrawal period.
    \item \textbf{UI Dataset}: This dataset consists of 28 participants, including 14 patients with PD and 14 healthy controls. EEG signals were collected only while the patients were on dopaminergic medication. 
\end{itemize}

\begin{figure*}[!t]
    \centering
    \includegraphics[width=0.8\textwidth]{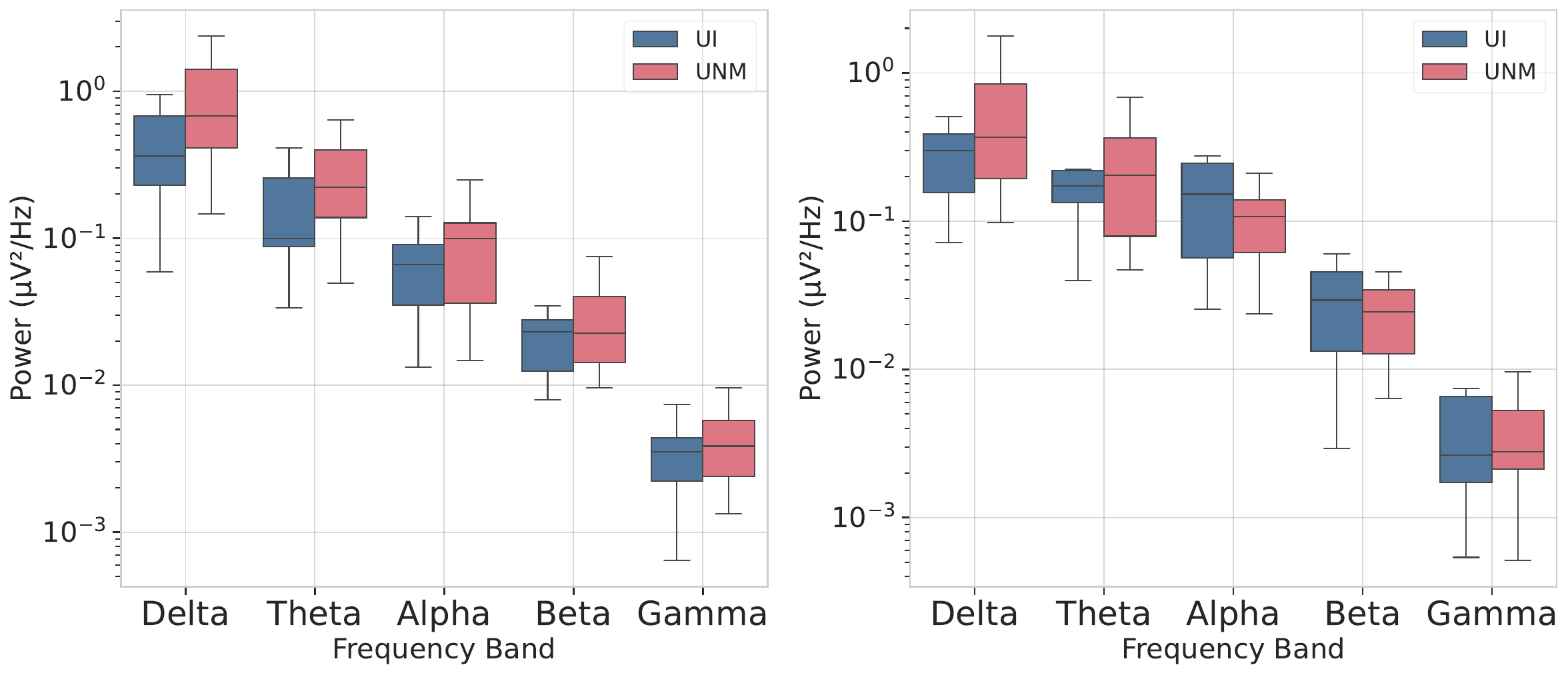}
    \caption{Power spectral density analysis across different datasets and classes: Comparison of frequency band-specific PSD between UI and UNM in EEG dataset. (Left) Healthy controls. (Right) Parkinson's disease. Blue and red denote the UI and UNM dataset, respectively. Note that the y-axis is shown on a logarithmic scale for better visualization.}
    \label{fig:PSDAnalysisEEG}
\end{figure*}

\begin{figure*}[!t]
    \centering
    \includegraphics[width=0.8\textwidth]{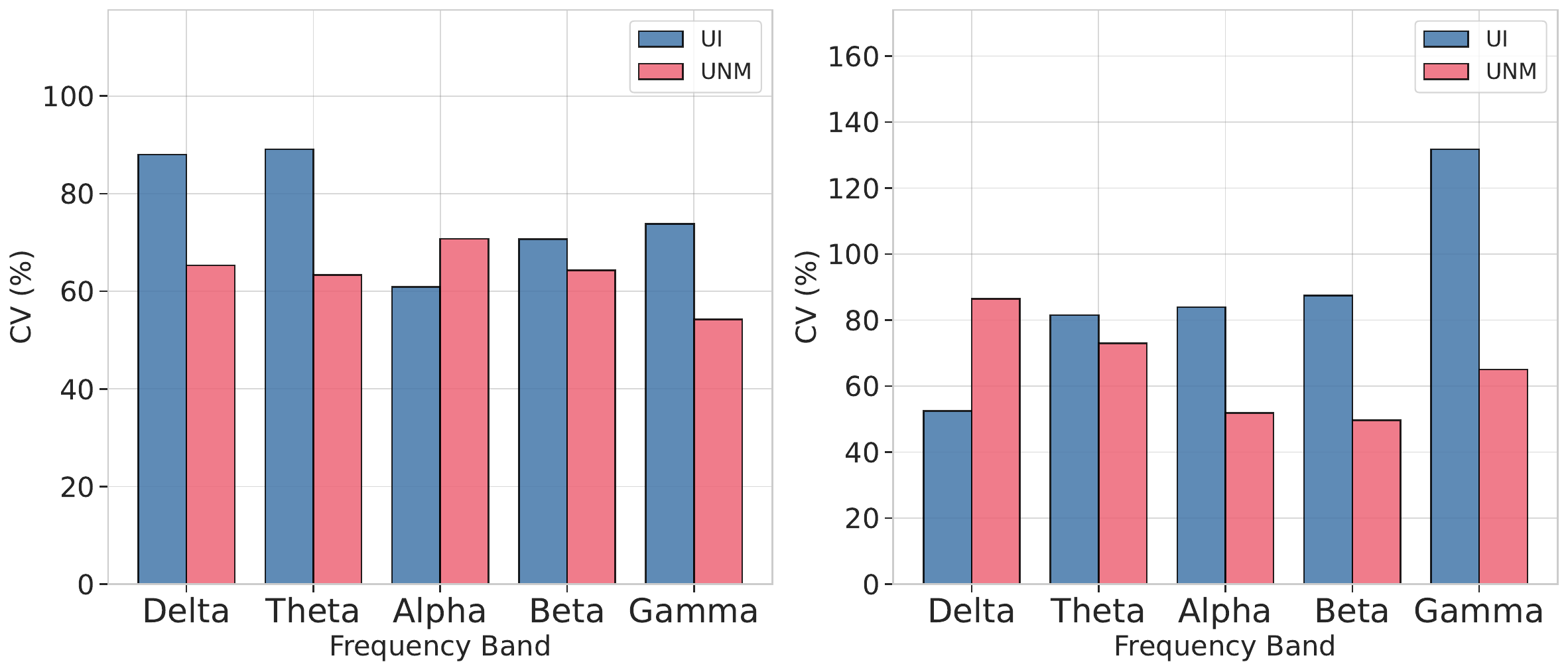}
    \caption{Inter-subject variability analysis using coefficient of variation (CV): Comparison of inter-subject variability between UI and UNM in EEG dataset. (Left) Healthy controls. (Right) Parkinson's disease. Blue and red denote the UI and UNM dataset, respectively.}
    \label{fig:PSDCVAnalysisEEG}
\end{figure*}

Both EEG datasets were recorded using a 64-channel BrainVision system at a sampling rate of 500 Hz. Detailed descriptions of the datasets are provided in the original publications \citep{cavanagh2018diminished, singh2020frontal}. In these datasets, preprocessing was performed following the procedures described in previous studies, and ASR was not applied \citep{sugden2023generalizable}. Electrodes that were not located at consistent positions across both datasets were excluded, resulting in 60 EEG channels used for analysis. Table~\ref{table:EEGSample} summarizes the number of subjects and samples in the UNM and UI datasets. Both datasets contain a relatively large number of subjects but a limited number of samples per subject. 

\subsubsection{Dataset distribution analysis}
\label{sec4.2.2}

Although the two datasets have similar average participant ages, a considerable domain gap exists due to differences in recording locations, institutions, subject composition, medication status, and experimental protocols \citep{karakas2023distinguishing}. Figure~\ref{fig:EEGPsdTsne} shows the t-SNE visualization of the PSD features, indicating a clear distributional discrepancy between the UNM and UI datasets. The silhouette score analysis (Fig.~\ref{fig:EEGSSHistogram}) further supported this observation, revealing clear subject-wise clustering in both datasets. In addition, the UI dataset exhibited stronger inter-subject separation than the UNM dataset.

Furthermore, we quantitatively analyzed the PSD distributions across frequency bands and the inter-subject variability within the UI and UNM datasets. During this analysis, extreme outliers were observed in the UNM data, particularly in the Delta and Theta bands. Therefore, we removed these outliers using the interquartile range ($\text{IQR}$) method, where values outside the range $[Q_1 - 1.5 \times \text{IQR},\; Q_3 + 1.5 \times \text{IQR}]$ for each frequency band were excluded.

As shown in Fig.~\ref{fig:PSDAnalysisEEG}, both the UNM and UI datasets exhibited higher PSD values in the low-frequency bands (i.e., Delta, Theta, Alpha) than in the high-frequency bands (i.e., Beta, Gamma). In both datasets, the PSD distributions differed between the PD and healthy control groups, showing distinct PSD patterns between the two populations. Notably, the UNM dataset showed overall higher PSD values in the low-frequency bands compared to the UI dataset. These distributional differences suggest a dataset-wise domain gap caused by variations in recording environments, institutions, and participant populations.

Figure~\ref{fig:PSDCVAnalysisEEG} shows the CV of the mean PSD values for each subject. In the UI dataset, the healthy control and PD groups exhibited distinct variability patterns: the healthy control group showed higher variability in the low-frequency bands, whereas the PD group showed greater variability in the high-frequency range. In contrast, the UNM dataset displayed uniformly high CV values across all frequency bands for both groups. These findings reveal large inter-subject variability within both datasets.

\section{Experiments}
\label{sec5}

\subsection{Baselines}
\label{sec5.1}

In this section, we compared the proposed method with the following six baseline models. All models were trained and evaluated on the same preprocessed data. For the ECoG dataset, we applied the same preprocessing pipeline---band-pass filtering, notch filtering, and ASR---to all recordings before feeding them into each model. For the EEG dataset, we followed the preprocessing protocol of \citet{wang2024dmmr} and used the same preprocessed data across all models.
\begin{itemize}
    \item \textbf{EEGNet} \citep{lawhern2018eegnet}: A compact CNN architecture widely used for EEG signal classification. It extracts spatio-temporal features using depthwise and separable convolutions. In this study, EEGNet serves as a baseline model without any domain generalization approaches.
    \item \textbf{DBConformer} \citep{wang2025dbconformer}: A dual-branch CNN-Transformer architecture that employs parallel temporal and spatial Conformer\- branches to simultaneously capture long-range temporal dependencies and inter-channel interactions in EEG signals. In this study, DBConformer was used as a baseline model without any domain generalization approaches.
    \item \textbf{TCFormer} \citep{altaheri2025temporal}: A CNN-Transformer architecture that integrates multi-kernel convolutions for multi-scale spatio-temporal feature extraction, combined with a grouped-query attention Transformer and a temporal convolutional network head for decoding. This model also serves as a baseline without domain generalization.
    \item \textbf{GroupDRO} \citep{kim2025explainable}: A domain generalization method based on distributionally robust optimization that minimizes the worst-case loss over pre-defined groups rather than the average loss.
    \item \textbf{DMMR} \citep{wang2024dmmr}: A domain generalization model that leverages self-supervised learning to obtain subject-invariant representations. During pretraining, a multi-decoder autoencoder is used to extract invariant features, with time steps shuffling-based noise injection to enhance robustness. In the fine-tuning phase, the pretrained encoder is integrated with a target classifier for downstream tasks.
    \item \textbf{EEG-DG} \citep{zhong2024eegdg}: A multi-source domain generalization framework that optimizes both marginal and conditional distributions across source domains to learn domain-invariant feature representations generalizable to unseen target subjects.
\end{itemize}

\begin{table*}[!t]
    \centering
    \begin{minipage}[t]{0.38\textwidth}
        \centering
        \caption{Hyperparameter values of $\lambda_{MI}$ and $\lambda_{GRL}$ for each experimental setting.}
        \label{table:lambdaHP}
        \sctablesetup
        \resizebox{\linewidth}{!}{%
        \begin{tabular}{@{}lllll@{}}
            \toprule
            Data type & \makecell[l]{Training\\dataset} & \makecell[l]{Test\\dataset} & $\lambda_{MI}$ & $\lambda_{GRL}$ \\
            \midrule
            \multirow{4}{*}{ECoG dataset}
             & Wireless & Wireless & 1.1120 & 0.7413 \\
             & Wired & Wired & 1.1120 & 0.5833 \\
             & Wireless & Wired & 4.4450 & 0.6667 \\
             & Wired & Wireless & 7.7780 & 0.8155 \\
            \midrule
            \multirow{2}{*}{EEG dataset}
             & UI & UNM & 0.417 & 4.445 \\
             & UNM & UI & 0.001 & 1.112 \\
            \bottomrule
        \end{tabular}
        }
    \end{minipage}
    \hfill
    \begin{minipage}[t]{0.58\textwidth}
        \centering
        \caption{Summary of experiments. This table presents the experimental settings used to evaluate model performance. Experiment 1 investigates cross-subject generalization within each ECoG data, whereas Experiment 2 investigates cross-cohort generalization under the compound domain shift between the wired and wireless cohorts.}
        \label{table:summary_exp}
        \sctablesetup
        \resizebox{\linewidth}{!}{%
        \begin{tabular}{@{}lllll@{}}
            \toprule
             & Data type & \makecell[l]{Training\\dataset} & \makecell[l]{Test\\dataset} & Goal \\
            \midrule
            \multirow{2}{*}{Experiment 1} & \multirow{2}{*}{ECoG} & Wireless & Wireless & \multirow{2}{*}{\makecell[l]{Model performance on unseen subjects\\(Cross-subject generalization)}} \\
             &  & Wired & Wired &  \\
            \midrule
            \multirow{2}{*}{Experiment 2} & \multirow{2}{*}{ECoG} & Wireless & Wired & \multirow{2}{*}{\makecell[l]{Model performance on unseen domains\\(Cross-cohort generalization)}} \\
             &  & Wired & Wireless &  \\
            \midrule
            \multirow{4}{*}{Experiment 3} & \multirow{4}{*}{ECoG} & Wireless & Wireless & \multirow{4}{*}{Ablation study} \\
             &  & Wired & Wired &  \\
             &  & Wireless & Wired &  \\
             &  & Wired & Wireless &  \\
            \midrule
            \multirow{2}{*}{Experiment 4} & \multirow{2}{*}{EEG} & UI & UNM & \multirow{2}{*}{\makecell[l]{Model performance on unseen domains\\(Cross-dataset generalization)}} \\
             &  & UNM & UI &  \\
            \bottomrule
        \end{tabular}
        }
    \end{minipage}
\end{table*}

\subsection{Evaluation metrics}
\label{sec5.2}
In this study, we evaluated binary classification performance using multiple metrics: accuracy, precision, and the F1-score. All metrics were computed using macro-averaging, where scores were calculated for each class and then averaged. Macro recall was not reported separately because, under the class-wise calculation used in this study, it was mathematically identical to macro accuracy. These metrics followed standard definitions commonly used for classification tasks.

\subsection{Implementation details}
\label{sec5.3}

\noindent\textbf{ASR Preprocessing Settings.} For the ECoG recordings, we used the ASR function implemented in EEGLAB's clean\_asr() module, with all parameters kept at their default settings without additional tuning.

\noindent\textbf{Parameter Settings.} For EEGNet, the temporal kernel length was set to half of the sampling rate, as suggested in the original paper (\textit{i.e.}, ECoG: 256~Hz, EEG: 250~Hz). The number of channels was set to match those available in each dataset. For the proposed method, we conducted a grid search over two hyperparameters, $\lambda_{MI}$ and $\lambda_{GRL}$, uniformly sampled within the range [0.001, 10] using 25 and 10 intervals, respectively. 
The hyperparameter combination with the highest validation macro accuracy was selected as the optimal setting. The optimal values for each training dataset were selected based on this search, and the final configurations are summarized in Table~\ref{table:lambdaHP}.
For ISBCS, the channel-swap probability was set to $p = 0.5$.

\noindent\textbf{Optimization Settings.} All models were trained using the Adam optimizer with an initial learning rate of 0.001, while other hyperparameters were kept at their default values. The learning rate was dynamically adjusted by a scheduler, which halved the learning rate when the validation accuracy did not improve by more than 0.001 for five consecutive epochs. To ensure convergence, each model was trained for at least 20 epochs, and early stopping with a patience of 10 was applied to prevent overfitting. 
All experiments were independently repeated using 10 different random seeds, and we report the mean and standard deviation of the results to characterize variability across repeated training runs. For the ECoG experiments, which included three independent rats in each modality group, we additionally performed subject-level paired analyses. For each rat, performance was first averaged across the same 10 matched random seeds to obtain a single subject-level score for each method. The paired performance difference was then calculated as SAF minus the corresponding baseline for the same rat. Across the three rats, we report the 95\% confidence interval for the mean paired difference, paired Cohen's effect size ($d_z$), and matched-pairs rank-biserial correlation ($r_{\mathrm{rb}}$). Individual rat-level paired differences are also reported to show the consistency of the observed effects across subjects.

\subsection{Summary of experiments}
\label{sec5.4}
Four experiments were conducted on the ECoG dataset as described below. Experiments 1 and 2 evaluated the effects of cross-subject and cross-cohort generalization within the ECoG dataset, respectively. Experiment~3 conducted an ablation study to evaluate the effectiveness and compatibility of the proposed ISBCS and DAL modules. Finally, Experiment 4 further validated the generalization capability of the proposed method using a publicly available EEG dataset, as summarized in Table~\ref{table:summary_exp}.

The predefined training, validation, and test sets were used consistently according to the domain-generalization setting of each experiment.

In Experiment 1, one of the three rats within each ECoG modality group was completely held out as the target subject. Only the training and validation sets of the remaining two rats were used for model training and selection, whereas the test set of the held-out subject was used exclusively for final evaluation. The training and validation sets of the held-out subject were not used at any stage of model training or selection. This leave-one-subject-out (LOSO) procedure was repeated until each rat had served once as the target subject.

In Experiment 2, one of the wireless and wired ECoG cohorts was completely held out as the target cohort. Only the training and validation sets of the other cohort were used for model training and selection, whereas the test sets of the held-out cohort were used exclusively for final evaluation. The training and validation sets of the held-out cohort were not used at any stage of model training or selection. The two evaluation directions were defined by setting the wired cohort as the target cohort for Wireless$\to$Wired and the wireless cohort as the target cohort for Wired$\to$Wireless.

Experiment 3 used the same training, validation, and test sets as Experiments 1 and 2 for the corresponding ablation settings.

In Experiment 4, one of the two public EEG datasets, UI and UNM, was completely held out as the target dataset. Only the training and validation sets of the other dataset were used for model training and selection, whereas the test set of the held-out dataset was used exclusively for final evaluation. The training and validation sets of the held-out dataset were not used at any stage of model training or selection. The two evaluation directions were defined by setting UNM as the target dataset for UI$\to$UNM and UI as the target dataset for UNM$\to$UI.

\begin{table*}[!t]
    \centering
    \caption{Domain generalization results on the ECoG dataset (Experiments 1 and 2). Here, the notation `A $\to$ B' denotes that each model was trained using dataset A and evaluated on dataset B.}
    \label{table:Experiment12}
    \sctablesetup
    \begin{tabular}{lllll}
        \toprule
        Experiment & Method & Accuracy (\%) & Precision (\%) & F1-score (\%) \\
        \midrule
        \multirow{7}{*}{\makecell[l]{Wireless$\to$Wireless\\(LOSO)}}
          & EEGNet & 57.89 $\pm$ 7.88 & 66.38 $\pm$ 12.16 & 50.94 $\pm$ 10.00 \\
          & DBConformer & 63.96 $\pm$ 5.81 & 64.56 $\pm$ 11.81 & 56.74 $\pm$ 6.91 \\
          & TCFormer & 59.55 $\pm$ 6.64 & 59.95 $\pm$ 8.17 & 51.57 $\pm$ 7.31 \\
          & GroupDRO & 55.93 $\pm$ 6.57 & 55.65 $\pm$ 8.50 & 47.12 $\pm$ 8.88 \\
          & DMMR & 58.46 $\pm$ 12.14 & 52.11 $\pm$ 21.65 & 51.00 $\pm$ 16.17 \\
          & EEG-DG & 60.25 $\pm$ 7.59 & 61.25 $\pm$ 7.76 & 50.37 $\pm$ 10.02 \\
        \cmidrule{2-5}
          & Ours (SAF) & \textbf{69.20 $\pm$ 9.23} & \textbf{73.21 $\pm$ 12.93} & \textbf{65.18 $\pm$ 12.38} \\
        \midrule
        \multirow{7}{*}{\makecell[l]{Wired$\to$Wired\\(LOSO)}}
          & EEGNet & 52.06 $\pm$ 2.54 & 40.87 $\pm$ 5.99 & 37.86 $\pm$ 3.56 \\
          & DBConformer & 52.84 $\pm$ 0.61 & 42.54 $\pm$ 0.14 & 39.68 $\pm$ 0.78 \\
          & TCFormer & 53.90 $\pm$ 3.79 & 39.96 $\pm$ 4.49 & 41.42 $\pm$ 4.36 \\
          & GroupDRO & 53.07 $\pm$ 2.98 & 35.94 $\pm$ 4.72 & 41.10 $\pm$ 4.06 \\
          & DMMR & 53.49 $\pm$ 2.81 & 51.87 $\pm$ 6.60 & 42.55 $\pm$ 6.57 \\
          & EEG-DG & 53.19 $\pm$ 1.60 & 39.65 $\pm$ 7.45 & 38.67 $\pm$ 3.48 \\
        \cmidrule{2-5}
          & Ours (SAF) & \textbf{58.32 $\pm$ 3.28} & \textbf{58.87 $\pm$ 6.09} & \textbf{48.02 $\pm$ 4.58} \\
        \midrule
        \multirow{7}{*}{Wireless$\to$Wired}
          & EEGNet & 50.24 $\pm$ 1.31 & 32.70 $\pm$ 4.57 & 37.61 $\pm$ 2.44 \\
          & DBConformer & 63.76 $\pm$ 1.40 & 48.89 $\pm$ 1.90 & 53.64 $\pm$ 1.71 \\
          & TCFormer & 62.67 $\pm$ 5.39 & 49.80 $\pm$ 1.32 & 53.78 $\pm$ 2.73 \\
          & GroupDRO & 57.34 $\pm$ 8.59 & 58.80 $\pm$ 18.73 & 50.82 $\pm$ 9.97 \\
          & DMMR & 57.31 $\pm$ 9.12 & 51.93 $\pm$ 15.11 & 50.93 $\pm$ 12.92 \\
          & EEG-DG & 58.97 $\pm$ 6.52 & \textbf{66.19 $\pm$ 16.11} & 52.56 $\pm$ 6.89 \\
        \cmidrule{2-5}
          & Ours (SAF) & \textbf{64.18 $\pm$ 4.64} & 59.83 $\pm$ 9.88 & \textbf{59.23 $\pm$ 7.71} \\
        \midrule
        \multirow{7}{*}{Wired$\to$Wireless}
          & EEGNet & 57.62 $\pm$ 7.52 & 66.03 $\pm$ 11.05 & 48.28 $\pm$ 11.45 \\
          & DBConformer & 47.34 $\pm$ 5.52 & 31.31 $\pm$ 8.11 & 34.31 $\pm$ 4.98 \\
          & TCFormer & 48.86 $\pm$ 4.34 & 44.94 $\pm$ 6.75 & 43.28 $\pm$ 6.77 \\
          & GroupDRO & 48.56 $\pm$ 2.13 & 24.13 $\pm$ 0.44 & 32.11 $\pm$ 0.96 \\
          & DMMR & 51.98 $\pm$ 3.94 & 53.55 $\pm$ 10.34 & 43.69 $\pm$ 5.42 \\
          & EEG-DG & 55.49 $\pm$ 4.05 & 61.42 $\pm$ 8.13 & 49.12 $\pm$ 8.45 \\
        \cmidrule{2-5}
          & Ours (SAF) & \textbf{69.90 $\pm$ 6.67} & \textbf{73.67 $\pm$ 6.27} & \textbf{67.19 $\pm$ 8.35} \\
        \bottomrule
    \end{tabular}
\end{table*}

\subsection{Results on domain generalization (ECoG dataset)}
\label{sec5.5}

Table~\ref{table:Experiment12} presents the domain generalization results on the ECoG dataset. Here, the notation `A $\to$ B' denotes that each model was trained using dataset A and evaluated on dataset B. The results of Experiment 1 show the cross-subject generalization performance of our proposed framework compared to baseline models on the ECoG dataset under both wireless and wired conditions. {The results indicate that the proposed framework achieves numerically higher performance on average than all baselines across all three evaluation metrics for unseen subjects. We observe that the classification performance under the wireless condition is generally higher than that under the wired condition.} As illustrated in Fig.~\ref{fig:PSDAnalysisMOCOP}, this may be due to the fact that the class-discriminative information is more distinct in the wireless condition than in the wired condition. For instance, looking at the $\delta$-band in Fig.~\ref{fig:PSDAnalysisMOCOP}, the PSD is clearly distinguishable before and after electrical stimulation under the wireless condition.

In Experiment 2, we evaluated the proposed framework in a cross-cohort generalization setting by training on one ECoG cohort and testing on the other. The wireless-to-wired and wired-to-wireless directions involve different animals as well as differences in acquisition system, recording duration, original sampling rate, and recording conditions. Accordingly, the results characterize generalization under a compound cross-cohort shift. {The proposed framework achieved the highest accuracy and F1-score across both transfer settings. Although EEG-DG achieved the highest precision in the wireless-to-wired setting, its mean accuracy and F1-score remained lower than those of the proposed framework.}

Subject-level paired comparisons between SAF and the six baselines across the four ECoG evaluation settings are summarized in Tables~\ref*{table:paired_wireless_loso}--\ref*{table:paired_wired_wireless}. Across the 72 comparisons comprising four evaluation settings, three metrics, and six baselines, paired Cohen's effect size ($d_z$) favored SAF in all comparisons, while the matched-pairs rank-biserial correlation ($r_{\mathrm{rb}}$) favored SAF in 67 comparisons. {These results indicate that, despite the limited number of subjects, SAF showed numerically positive performance differences relative to the baseline models across most comparisons. Although statistical significance was not established for many comparisons, as their 95\% confidence intervals included zero, SAF showed numerically higher performance on average under the evaluated ECoG conditions.}


\begin{table*}[!htbp]
    \centering
    \caption{Subject-level paired comparisons between SAF and six methods in the Wireless$\to$Wireless (LOSO) setting.}
    \label{table:paired_wireless_loso}
    \sctablesetup
    \begin{tabular}{llllllll}
        \toprule
        Metric & Method & $\Delta$ Rat1 & $\Delta$ Rat2 & $\Delta$ Rat3 & 95\% CI of mean $\Delta$ & $d_z$ & $r_{\mathrm{rb}}$ \\
        \midrule
        \multirow{6}{*}{Accuracy (\%)} & EEGNet & $+18.01$ & $+3.13$ & $+12.19$ & $[-7.51, +29.73]$ & $+1.48$ & $+1.00$ \\
         & DBConformer & $+16.69$ & $-22.11$ & $+20.22$ & $[-53.41, +63.28]$ & $+0.21$ & $0.00$ \\
         & TCFormer & $+14.55$ & $-7.02$ & $+21.11$ & $[-27.02, +46.11]$ & $+0.65$ & $+0.67$ \\
         & GroupDRO & $+12.51$ & $+5.00$ & $+22.67$ & $[-8.63, +35.42]$ & $+1.51$ & $+1.00$ \\
         & DMMR & $+11.73$ & $+19.80$ & $+22.09$ & $[+4.36, +31.39]$ & $+3.29$ & $+1.00$ \\
         & EEG-DG & $+7.99$ & $-3.07$ & $+21.11$ & $[-21.39, +38.74]$ & $+0.72$ & $+0.67$ \\
        \midrule
        \multirow{6}{*}{Precision (\%)} & EEGNet & $+21.98$ & $-2.25$ & $+2.64$ & $[-24.37, +39.28]$ & $+0.58$ & $+0.67$ \\
         & DBConformer & $+19.67$ & $-22.78$ & $+29.47$ & $[-60.21, +77.78]$ & $+0.32$ & $+0.33$ \\
         & TCFormer & $+7.16$ & $-14.76$ & $+49.37$ & $[-67.04, +94.89]$ & $+0.43$ & $+0.33$ \\
         & GroupDRO & $+8.69$ & $-2.62$ & $+48.21$ & $[-48.21, +84.39]$ & $+0.68$ & $+0.67$ \\
         & DMMR & $+26.78$ & $+37.90$ & $+37.28$ & $[+18.46, +49.51]$ & $+5.44$ & $+1.00$ \\
         & EEG-DG & $-3.75$ & $-7.93$ & $+49.37$ & $[-66.78, +91.91]$ & $+0.39$ & $0.00$ \\
        \midrule
        \multirow{6}{*}{F1-score (\%)} & EEGNet & $+21.90$ & $+3.71$ & $+18.11$ & $[-9.27, +38.41]$ & $+1.52$ & $+1.00$ \\
         & DBConformer & $+18.56$ & $-26.15$ & $+30.69$ & $[-66.66, +82.06]$ & $+0.26$ & $+0.33$ \\
         & TCFormer & $+17.83$ & $-9.56$ & $+32.33$ & $[-39.31, +66.38]$ & $+0.64$ & $+0.67$ \\
         & GroupDRO & $+15.60$ & $+6.96$ & $+32.65$ & $[-14.07, +50.88]$ & $+1.41$ & $+1.00$ \\
         & DMMR & $+16.62$ & $+26.93$ & $+26.52$ & $[+8.86, +37.86]$ & $+4.00$ & $+1.00$ \\
         & EEG-DG & $+13.69$ & $-1.15$ & $+32.33$ & $[-26.71, +56.62]$ & $+0.89$ & $+0.67$ \\
        \bottomrule
    \end{tabular}
    \par\smallskip
    \begin{minipage}{\textwidth}
        \footnotesize
        \textit{Note.} $\Delta$ denotes the paired performance difference (SAF $-$ baseline) in percentage points, and the 95\% CI corresponds to the mean paired difference across the three rats. $d_z$ denotes paired Cohen\textquotesingle s effect size and $r_{\mathrm{rb}}$ denotes the matched-pairs rank-biserial correlation. Positive values favor SAF.
    \end{minipage}
\end{table*}

\begin{table*}[!htbp]
    \centering
    \caption{Subject-level paired comparisons between SAF and six methods in the Wired$\to$Wired (LOSO) setting.}
    \label{table:paired_wired_loso}
    \sctablesetup
    \begin{tabular}{llllllll}
        \toprule
        Metric & Method & $\Delta$ Rat1 & $\Delta$ Rat2 & $\Delta$ Rat3 & 95\% CI of mean $\Delta$ & $d_z$ & $r_{\mathrm{rb}}$ \\
        \midrule
        \multirow{6}{*}{Accuracy (\%)} & EEGNet & $+3.57$ & $+1.84$ & $+9.98$ & $[-5.53, +15.79]$ & $+1.20$ & $+1.00$ \\
         & DBConformer & $+1.50$ & $+1.84$ & $+9.78$ & $[-7.26, +16.01]$ & $+0.93$ & $+1.00$ \\
         & TCFormer & $-0.17$ & $+1.84$ & $+10.20$ & $[-9.70, +17.62]$ & $+0.72$ & $+0.67$ \\
         & GroupDRO & $+2.05$ & $+1.84$ & $+9.78$ & $[-6.67, +15.79]$ & $+1.01$ & $+1.00$ \\
         & DMMR & $+3.85$ & $+0.07$ & $+9.77$ & $[-7.59, +16.71]$ & $+0.93$ & $+1.00$ \\
         & EEG-DG & $-0.28$ & $+1.84$ & $+9.60$ & $[-9.20, +16.64]$ & $+0.72$ & $+0.67$ \\
        \midrule
        \multirow{6}{*}{Precision (\%)} & EEGNet & $+7.90$ & $+13.47$ & $+29.19$ & $[-10.57, +44.28]$ & $+1.53$ & $+1.00$ \\
         & DBConformer & $-0.92$ & $+13.47$ & $+33.08$ & $[-27.18, +57.60]$ & $+0.89$ & $+0.67$ \\
         & TCFormer & $+9.57$ & $+13.47$ & $+31.52$ & $[-10.90, +47.28]$ & $+1.55$ & $+1.00$ \\
         & GroupDRO & $+18.67$ & $+13.47$ & $+33.08$ & $[-3.49, +46.97]$ & $+2.14$ & $+1.00$ \\
         & DMMR & $+12.56$ & $-20.95$ & $+30.97$ & $[-57.86, +72.91]$ & $+0.29$ & $+0.33$ \\
         & EEG-DG & $+5.91$ & $+13.47$ & $+37.90$ & $[-22.43, +60.62]$ & $+1.14$ & $+1.00$ \\
        \midrule
        \multirow{6}{*}{F1-score (\%)} & EEGNet & $+2.47$ & $+4.59$ & $+20.62$ & $[-15.42, +33.88]$ & $+0.93$ & $+1.00$ \\
         & DBConformer & $+2.14$ & $+4.59$ & $+16.49$ & $[-11.32, +26.80]$ & $+1.01$ & $+1.00$ \\
         & TCFormer & $-2.94$ & $+4.59$ & $+16.04$ & $[-17.85, +29.64]$ & $+0.62$ & $+0.67$ \\
         & GroupDRO & $-3.06$ & $+4.59$ & $+16.49$ & $[-18.46, +30.47]$ & $+0.61$ & $+0.67$ \\
         & DMMR & $+1.62$ & $-2.26$ & $+16.78$ & $[-19.62, +30.38]$ & $+0.53$ & $+0.33$ \\
         & EEG-DG & $-1.09$ & $+4.59$ & $+21.63$ & $[-21.00, +37.76]$ & $+0.71$ & $+0.67$ \\
        \bottomrule
    \end{tabular}
    \par\smallskip
    \begin{minipage}{\textwidth}
        \footnotesize
        \textit{Note.} $\Delta$ denotes the paired performance difference (SAF $-$ baseline) in percentage points, and the 95\% CI corresponds to the mean paired difference across the three rats. $d_z$ denotes paired Cohen\textquotesingle s effect size and $r_{\mathrm{rb}}$ denotes the matched-pairs rank-biserial correlation. Positive values favor SAF.
    \end{minipage}
\end{table*}

\begin{table*}[!htbp]
    \centering
    \caption{Subject-level paired comparisons between SAF and six methods in the Wireless$\to$Wired setting.}
    \label{table:paired_wireless_wired}
    \sctablesetup
    \begin{tabular}{llllllll}
        \toprule
        Metric & Method & $\Delta$ Rat1 & $\Delta$ Rat2 & $\Delta$ Rat3 & 95\% CI of mean $\Delta$ & $d_z$ & $r_{\mathrm{rb}}$ \\
        \midrule
        \multirow{6}{*}{Accuracy (\%)} & EEGNet & $0.00$ & $0.00$ & $+41.35$ & $[-45.53, +73.10]$ & $+0.58$ & $+1.00$ \\
         & DBConformer & $0.00$ & $0.00$ & $+1.07$ & $[-1.17, +1.89]$ & $+0.58$ & $+1.00$ \\
         & TCFormer & $+1.62$ & $0.00$ & $+2.33$ & $[-1.65, +4.28]$ & $+1.10$ & $+1.00$ \\
         & GroupDRO & $+0.40$ & $0.00$ & $+14.89$ & $[-15.97, +26.17]$ & $+0.60$ & $+1.00$ \\
         & DMMR & $-2.02$ & $+0.43$ & $+36.22$ & $[-41.63, +64.71]$ & $+0.54$ & $+0.33$ \\
         & EEG-DG & $+1.62$ & $0.00$ & $+7.76$ & $[-7.04, +13.29]$ & $+0.76$ & $+1.00$ \\
        \midrule
        \multirow{6}{*}{Precision (\%)} & EEGNet & $+0.03$ & $0.00$ & $+49.08$ & $[-54.01, +86.74]$ & $+0.58$ & $+1.00$ \\
         & DBConformer & $+0.03$ & $0.00$ & $+1.98$ & $[-2.16, +3.50]$ & $+0.59$ & $+1.00$ \\
         & TCFormer & $+0.51$ & $0.00$ & $+1.32$ & $[-1.05, +2.27]$ & $+0.91$ & $+1.00$ \\
         & GroupDRO & $+0.13$ & $0.00$ & $+17.69$ & $[-19.34, +31.22]$ & $+0.58$ & $+1.00$ \\
         & DMMR & $-18.30$ & $-12.32$ & $+37.02$ & $[-73.28, +77.54]$ & $+0.07$ & $0.00$ \\
         & EEG-DG & $-0.34$ & $0.00$ & $+3.87$ & $[-4.64, +6.99]$ & $+0.50$ & $+0.33$ \\
        \midrule
        \multirow{6}{*}{F1-score (\%)} & EEGNet & $+0.02$ & $0.00$ & $+51.30$ & $[-56.45, +90.67]$ & $+0.58$ & $+1.00$ \\
         & DBConformer & $+0.02$ & $0.00$ & $+1.28$ & $[-1.39, +2.26]$ & $+0.59$ & $+1.00$ \\
         & TCFormer & $+0.83$ & $0.00$ & $+2.39$ & $[-1.95, +4.09]$ & $+0.88$ & $+1.00$ \\
         & GroupDRO & $+0.21$ & $0.00$ & $+18.83$ & $[-20.51, +33.20]$ & $+0.59$ & $+1.00$ \\
         & DMMR & $-8.26$ & $-6.02$ & $+38.06$ & $[-56.96, +72.82]$ & $+0.30$ & $0.00$ \\
         & EEG-DG & $+0.32$ & $0.00$ & $+10.49$ & $[-11.22, +18.42]$ & $+0.60$ & $+1.00$ \\
        \bottomrule
    \end{tabular}
    \par\smallskip
    \begin{minipage}{\textwidth}
        \footnotesize
        \textit{Note.} $\Delta$ denotes the paired performance difference (SAF $-$ baseline) in percentage points, and the 95\% CI corresponds to the mean paired difference across the three rats. $d_z$ denotes paired Cohen\textquotesingle s effect size and $r_{\mathrm{rb}}$ denotes the matched-pairs rank-biserial correlation. Positive values favor SAF.
    \end{minipage}
\end{table*}

\begin{table*}[!htbp]
    \centering
    \caption{Subject-level paired comparisons between SAF and six methods in the Wired$\to$Wireless setting.}
    \label{table:paired_wired_wireless}
    \sctablesetup
    \begin{tabular}{llllllll}
        \toprule
        Metric & Method & $\Delta$ Rat1 & $\Delta$ Rat2 & $\Delta$ Rat3 & 95\% CI of mean $\Delta$ & $d_z$ & $r_{\mathrm{rb}}$ \\
        \midrule
        \multirow{6}{*}{Accuracy (\%)} & EEGNet & $-6.43$ & $+37.46$ & $+2.55$ & $[-46.40, +68.79]$ & $+0.48$ & $+0.33$ \\
         & DBConformer & $+2.57$ & $+39.11$ & $+14.51$ & $[-27.56, +65.02]$ & $+1.01$ & $+1.00$ \\
         & TCFormer & $-4.84$ & $+41.16$ & $+18.64$ & $[-38.83, +75.47]$ & $+0.80$ & $+0.67$ \\
         & GroupDRO & $+6.37$ & $+36.90$ & $+12.71$ & $[-21.36, +58.69]$ & $+1.16$ & $+1.00$ \\
         & DMMR & $-7.51$ & $+32.66$ & $+18.34$ & $[-36.09, +65.08]$ & $+0.71$ & $+0.67$ \\
         & EEG-DG & $-7.29$ & $+33.50$ & $+11.93$ & $[-37.98, +63.40]$ & $+0.62$ & $+0.67$ \\
        \midrule
        \multirow{6}{*}{Precision (\%)} & EEGNet & $-14.74$ & $+44.31$ & $-4.23$ & $[-69.81, +86.70]$ & $+0.27$ & $0.00$ \\
         & DBConformer & $+21.95$ & $+63.60$ & $+40.64$ & $[-9.75, +93.88]$ & $+2.02$ & $+1.00$ \\
         & TCFormer & $+10.01$ & $+45.94$ & $+31.30$ & $[-15.80, +73.97]$ & $+1.61$ & $+1.00$ \\
         & GroupDRO & $+44.58$ & $+64.26$ & $+40.13$ & $[+17.75, +81.56]$ & $+3.87$ & $+1.00$ \\
         & DMMR & $-1.25$ & $+34.83$ & $+29.23$ & $[-27.29, +69.16]$ & $+1.08$ & $+0.67$ \\
         & EEG-DG & $-8.35$ & $+31.03$ & $+19.97$ & $[-36.24, +64.67]$ & $+0.70$ & $+0.67$ \\
        \midrule
        \multirow{6}{*}{F1-score (\%)} & EEGNet & $-7.92$ & $+50.57$ & $+1.75$ & $[-63.08, +92.68]$ & $+0.47$ & $+0.33$ \\
         & DBConformer & $+4.51$ & $+54.37$ & $+22.38$ & $[-35.67, +89.84]$ & $+1.07$ & $+1.00$ \\
         & TCFormer & $-9.27$ & $+46.47$ & $+20.11$ & $[-50.16, +88.37]$ & $+0.69$ & $+0.67$ \\
         & GroupDRO & $+13.88$ & $+53.43$ & $+21.39$ & $[-22.61, +81.74]$ & $+1.41$ & $+1.00$ \\
         & DMMR & $-13.15$ & $+42.82$ & $+20.92$ & $[-53.21, +86.93]$ & $+0.60$ & $+0.67$ \\
         & EEG-DG & $-11.48$ & $+41.83$ & $+12.73$ & $[-51.95, +80.67]$ & $+0.54$ & $+0.67$ \\
        \bottomrule
    \end{tabular}
    \par\smallskip
    \begin{minipage}{\textwidth}
        \footnotesize
        \textit{Note.} $\Delta$ denotes the paired performance difference (SAF $-$ baseline) in percentage points, and the 95\% CI corresponds to the mean paired difference across the three rats. $d_z$ denotes paired Cohen\textquotesingle s effect size and $r_{\mathrm{rb}}$ denotes the matched-pairs rank-biserial correlation. Positive values favor SAF.
    \end{minipage}
\end{table*}

\FloatBarrier

\subsection{Results on ablation study}
\label{sec5.6}

\begin{table*}[!t]
    \centering
    \caption{Ablation study results (Experiment 3). Here, the notation `A $\to$ B' denotes that each model was trained using dataset A and evaluated on dataset B. EEGNet serves as the baseline, and `+DAL' and `+ISBCS' indicate the addition of domain-adversarial learning and inter-subject balanced channel swap, respectively.}
    \label{table:Experiment3}
    \sctablesetup
    \begin{tabular}{lllll}
        \toprule
        Experiment & Method & Accuracy (\%) & Precision (\%) & F1-score (\%) \\
        \midrule
        \multirow{4}{*}{\makecell[l]{Wireless$\to$Wireless\\(LOSO)}}
          & EEGNet & 57.89 $\pm$ 7.88 & 66.38 $\pm$ 12.16 & 50.94 $\pm$ 10.00 \\
          & EEGNet+DAL & 53.72 $\pm$ 5.98 & 57.49 $\pm$ 19.97 & 43.49 $\pm$ 9.65 \\
          & EEGNet+ISBCS & 60.58 $\pm$ 6.85 & 59.74 $\pm$ 13.62 & 49.15 $\pm$ 11.82 \\
        \cmidrule{2-5}
          & Ours (SAF) & \textbf{69.20 $\pm$ 9.23} & \textbf{73.21 $\pm$ 12.93} & \textbf{65.18 $\pm$ 12.38} \\
        \midrule
        \multirow{4}{*}{\makecell[l]{Wired$\to$Wired\\(LOSO)}}
          & EEGNet & 52.06 $\pm$ 2.54 & 40.87 $\pm$ 5.99 & 37.86 $\pm$ 3.56 \\
          & EEGNet+DAL & 55.84 $\pm$ 13.58 & 49.05 $\pm$ 20.86 & 45.26 $\pm$ 15.61 \\
          & EEGNet+ISBCS & 53.79 $\pm$ 3.26 & 43.87 $\pm$ 10.57 & 41.84 $\pm$ 1.53 \\
        \cmidrule{2-5}
          & Ours (SAF) & \textbf{58.32 $\pm$ 3.28} & \textbf{58.87 $\pm$ 6.09} & \textbf{48.02 $\pm$ 4.58} \\
        \midrule
        \multirow{4}{*}{Wireless$\to$Wired}
          & EEGNet & 50.24 $\pm$ 1.31 & 32.70 $\pm$ 4.57 & 37.61 $\pm$ 2.44 \\
          & EEGNet+DAL & 50.00 $\pm$ 0.39 & 41.93 $\pm$ 12.28 & 37.14 $\pm$ 1.64 \\
          & EEGNet+ISBCS & 51.11 $\pm$ 1.22 & 45.73 $\pm$ 9.67 & 38.03 $\pm$ 2.66 \\
        \cmidrule{2-5}
          & Ours (SAF) & \textbf{64.18 $\pm$ 4.64} & \textbf{59.83 $\pm$ 9.88} & \textbf{59.23 $\pm$ 7.71} \\
        \midrule
        \multirow{4}{*}{Wired$\to$Wireless}
          & EEGNet & 57.62 $\pm$ 7.52 & 66.03 $\pm$ 11.05 & 48.28 $\pm$ 11.45 \\
          & EEGNet+DAL & 65.68 $\pm$ 10.68 & 63.12 $\pm$ 15.69 & 55.26 $\pm$ 15.75 \\
          & EEGNet+ISBCS & 63.73 $\pm$ 4.11 & 65.70 $\pm$ 2.92 & 57.02 $\pm$ 5.77 \\
        \cmidrule{2-5}
          & Ours (SAF) & \textbf{69.90 $\pm$ 6.67} & \textbf{73.67 $\pm$ 6.27} & \textbf{67.19 $\pm$ 8.35} \\
        \bottomrule
    \end{tabular}
\end{table*}

Table~\ref{table:Experiment3} presents the results of the ablation study to validate the contributions of ISBCS and DAL. Applying either component alone improved performance across all settings. When trained on wired data, the DAL-only model achieved higher accuracy than the ISBCS-only model, whereas the opposite trend was observed when trained on wireless data. To better understand the impact of ISBCS on the feature space, we analyzed its effect using t-SNE visualization. These results suggest that the two components respond differently to the factors constituting the compound domain shift, including animal-cohort differences and acquisition-related factors. Combining ISBCS and DAL achieved the highest performance across all settings, demonstrating that the two components work in a complementary manner.

\begin{figure}[!t]
    \centering
    \includegraphics[width=0.7\linewidth]{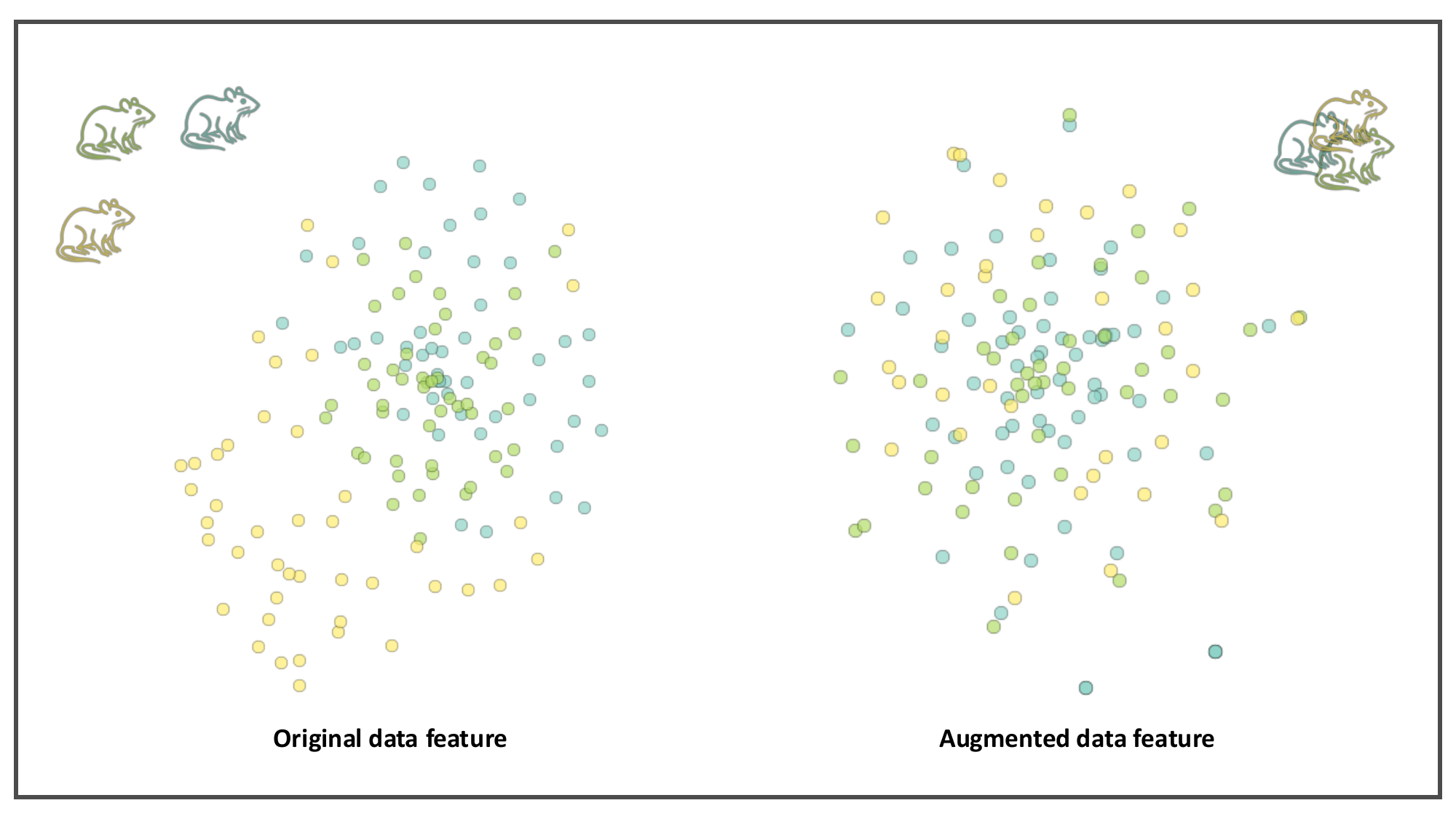}
    \caption{T-SNE visualization of the original and ISBCS-augmented data features. Each dot represents a sample in the feature space, and different colors correspond to different subjects. }
    \label{fig:ISBCSTsne}
\end{figure}

To better understand the impact of ISBCS on the feature space, we analyzed its effect using t-SNE visualization. Figure~\ref{fig:ISBCSTsne} visualizes the feature distributions extracted by the proposed framework from two types of inputs: the original data and the ISBCS-augmented data using t-SNE visualization. For the original data, features from different subjects are relatively well separated in the t-SNE space. After applying ISBCS, however, the features become less distinguishable across subjects, indicating that ISBCS effectively mitigates subject-specific characteristics through cross-subject channel exchange.
To quantitatively assess the degree of inter-subject separability, we computed  the F-statistic, which measures the ratio of between-subject variance to  within-subject variance. For the original data, the F-statistic was 24.48, indicating significant subject-specific differences. After applying ISBCS augmentation, the F-statistic dramatically decreased to 0.99, which indicates negligible between-subject differences. This reduction provides statistical evidence that ISBCS effectively reduces inter-subject variability.

\subsection{Results on domain generalization (public EEG dataset)}
\label{sec5.7}

\begin{table*}[!t]
    \centering
    \caption{Domain generalization results on EEG data (Experiment 4). Here, the notation `A $\to$ B' denotes that each model was trained using dataset A and evaluated on dataset B.}
    \label{table:Experiment4}
    \sctablesetup
    \begin{tabular}{lllll}
        \toprule
        Experiment & Method & Accuracy (\%) & Precision (\%) & F1-score (\%) \\
        \midrule
        \multirow{7}{*}{UI$\to$UNM}
          & EEGNet & 66.84 $\pm$ 4.38 & 71.55 $\pm$ 3.88 & 63.91 $\pm$ 5.70 \\
          & DBConformer & 57.63 $\pm$ 1.70 & 61.95 $\pm$ 2.94 & 52.72 $\pm$ 2.77 \\
          & TCFormer & 61.05 $\pm$ 1.82 & 62.37 $\pm$ 1.90 & 59.61 $\pm$ 2.62 \\
          & GroupDRO & 58.87 $\pm$ 1.58 & 60.88 $\pm$ 1.35 & 56.19 $\pm$ 2.73 \\
          & DMMR & 51.09 $\pm$ 4.26 & 51.06 $\pm$ 4.51 & 50.36 $\pm$ 4.69 \\
          & EEG-DG & 55.66 $\pm$ 4.59 & 55.21 $\pm$ 6.39 & 52.85 $\pm$ 8.70 \\
        \cmidrule{2-5}
          & Ours (SAF) & \textbf{73.34 $\pm$ 7.81} & \textbf{74.84 $\pm$ 6.40} & \textbf{72.36 $\pm$ 9.17} \\
        \midrule
        \multirow{7}{*}{UNM$\to$UI}
          & EEGNet & 75.04 $\pm$ 4.51 & 75.70 $\pm$ 4.78 & 74.85 $\pm$ 4.52 \\
          & DBConformer & 59.61 $\pm$ 10.56 & 59.65 $\pm$ 10.62 & 59.55 $\pm$ 10.60 \\
          & TCFormer & 52.01 $\pm$ 7.82 & 52.03 $\pm$ 8.25 & 51.55 $\pm$ 8.03 \\
          & GroupDRO & 56.38 $\pm$ 5.15 & 56.42 $\pm$ 5.19 & 56.28 $\pm$ 5.21 \\
          & DMMR & 51.20 $\pm$ 5.41 & 51.78 $\pm$ 6.60 & 48.95 $\pm$ 5.39 \\
          & EEG-DG & 56.00 $\pm$ 5.47 & 61.40 $\pm$ 14.39 & 47.99 $\pm$ 11.47 \\
        \cmidrule{2-5}
          & Ours (SAF) & \textbf{82.19 $\pm$ 4.92} & \textbf{82.33 $\pm$ 4.83} & \textbf{82.15 $\pm$ 4.95} \\
        \bottomrule
    \end{tabular}
\end{table*}

For the experiment 4, we evaluated the proposed framework on publicly available EEG datasets to examine its cross-dataset generalization. This setting is particularly challenging due to compounded domain differences across subjects, recording modalities, institutions, medication states, and environments. 
Table~\ref{table:Experiment4} presents the cross-dataset evaluation results between the UI and UNM datasets. The proposed method consistently outperformed the baselines across all experimental settings. These results indicate that the proposed method exhibits strong generalization and transferability across diverse and complex domain variations. 

When the model was trained on the UNM dataset and tested on the UI dataset, it yielded higher classification performance than in the reverse direction. This may be due to the fact that the UNM dataset contains a larger number of subjects and samples than the UI dataset.\citep{cavanagh2018diminished, singh2020frontal}. This broader coverage may help the model capture greater inter-subject variability, although the performance gap may also reflect target-domain separability and dataset-specific recording characteristics.

\FloatBarrier
\subsection{Generalization to different backbones}
\label{sec:backbone_generalization}
To further evaluate the applicability of SAF across different backbone architectures, we additionally applied the proposed framework to SecNet \citep{liang2025secnet} and FBMSNet \citep{liu2023fbmsnet}, two recent EEG decoding models with different feature-extraction structures. Each backbone was evaluated both with and without SAF using identical data partitions and training settings. Specifically, SecNet+SAF and FBMSNet+SAF denote SecNet and FBMSNet integrated with the ISBCS and DAL components of SAF, respectively.

Applying SAF to SecNet and FBMSNet generally improved generalization performance across the evaluated settings. Across 36 comparisons comprising six evaluation settings, three metrics, and two additional backbones, the SAF-augmented models achieved higher performance than their corresponding original backbones in 30 comparisons. In particular, performance improved in all comparisons on the public EEG datasets. {Detailed performance results are provided in Supplementary Tables S2 and S3, and subject-level paired analyses for the ECoG experiments are provided in Supplementary Tables S4--S7.} These results indicate that the performance gains achieved by SAF are not specific to EEGNet and support the applicability of the proposed framework to recent EEG decoding backbones with different feature-extraction structures.

\FloatBarrier

\section{Discussion}
\label{sec6}

\noindent\textbf{Key findings.}
Previous ECoG-based classification models have mainly focused on performance optimization rather than robustness to domain shifts. To address this limitation, we present a domain generalization framework for ECoG-based classification of neural states before and after electrical stimulation, which enables learning of domain-invariant representations. The proposed method shows strong domain generalization capability not only on ECoG data but also on EEG signals, indicating its effectiveness across different types of brain signals.

This study aims to reduce various domain discrepancies inherent in brain-signal data, including inter-subject physiological variability, sensor characteristics, and differences in recording environments. To this end, we combine the pre-processing augmentation method ISBCS with the in-processing DAL approach.

To evaluate the effectiveness of the two methods, we conducted extensive experiments on the individual contributions of ISBCS and DAL. In the wireless dataset, which exhibits large inter-subject variability, ISBCS contributed more significantly than DAL. Conversely, in the wired dataset with relatively small variability, the effect of ISBCS was more limited. These results suggest that an appropriate generalization strategy depends on the characteristics of the data and that the two components of the proposed framework work in a complementary manner, minimizing classification degradation while enhancing generalization capability.

{
\noindent\textbf{Limitations and future work.}
This study focused on domain generalization based on brain signals for classifying neural states before and after electrical stimulation. Future work will extend the proposed framework to broader BCI applications, including emotion recognition and neurological disorder analysis, thereby broadening its applicability to general cognitive and clinical research using brain signals.

Each ECoG cohort included only three rats as biological replicates, limiting the statistical power of the subject-level comparisons. Repeated training with 10 random seeds characterized variability across training runs but did not increase the number of biological replicates. Therefore, the statistical significance of the observed differences could not be reliably established.

Motor threshold testing provided functional confirmation that the electrode array was targeting the motor cortical region. However, the standardized electrode implantation procedure and motor threshold testing do not provide quantitative post-mortem registration of each individual electrode channel and therefore cannot establish exact channel-to-channel anatomical correspondence across animals.
}

Lastly, the proposed method involves two key hyperparameters, $\lambda_{MI}$ and $\lambda_{GRL}$, which significantly affect performance within DAL. Such hyperparameter sensitivity has also been observed in previous adversarial learning and domain generalization studies \citep{ganin2016domain, yang2020towards}. Future work will investigate strategies for setting appropriate initial values of $\lambda_{MI}$ and $\lambda_{GRL}$ or treating them as learnable parameters that can be dynamically optimized during training.

\section{Conclusion}
\label{sec7}
This study addressed the domain generalization problem in ECoG-based classification of neural states before and after electrical stimulation by constructing the first publicly available ECoG benchmark dataset and proposing a method for learning domain-invariant representations. {The proposed method combines ISBCS and DAL in a complementary manner, achieving numerically higher accuracy and F1-score on average than the baseline methods across various domain-shift scenarios, such as cross-subject and cross-cohort evaluations.} These results demonstrate that the proposed method effectively learns domain-invariant features under diverse domain variations. Furthermore, cross-dataset evaluations on public EEG benchmarks (i.e., UI and UNM) showed that the proposed framework is applicable to ECoG and EEG signals. This suggests its potential as a unified framework for electrophysiological signal analysis. Additional evaluations with SecNet and FBMSNet showed that the performance gains achieved by SAF were not specific to EEGNet, supporting the applicability of the proposed framework to EEG decoding backbones with different feature-extraction structures.

To facilitate further research in this direction, we have publicly released the constructed ECoG benchmark and source code. We expect that future work will advance this method by extending it to other neurological disorders, incorporating state-of-the-art architectures, and developing effective strategies for optimizing key hyperparameters.

\section*{CRediT authorship contribution statement}
\textbf{Seongwon Jin:} Writing - original draft, Software, Visualization, Validation, Data curation, Conceptualization. \textbf{Hanseul Choi:} Writing - original draft, Visualization, Data curation, Conceptualization. \textbf{Gaeun Kim:} Writing review \& editing, \textbf{Sunggu Yang:} Resources. \textbf{Sungho Park:} Writing review \& editing, Supervision, Conceptualization. \textbf{Jibum Kim:} Funding acquisition, Writing review \& editing, Project administration, Supervision, Conceptualization.

\section*{Declaration of competing interest}
The authors declare that they have no known competing financial interests or personal relationships that could have appeared to influence the work reported in this paper.

\section*{Funding}
This work was supported by the \aqrevision{National Research Foundation of Korea (NRF)} grant funded by the Korea government (\aqrevision{MSIT}) (\aqrevision{RS-2024-00346364}). This work was also supported by research funding from the Incheon National University Research Grant in 2024.

\section*{Data and code availability}
{
The ECoG dataset used in this study is publicly available on Zenodo at \url{https://zenodo.org/records/23012653}, whereas the source code is available on GitHub at \url{https://github.com/jinwork00/swap-adversarial-framework}.
The Zenodo repository includes a README file describing the dataset structure, sample counts, channel identities, and channel ordering.
}

\bibliographystyle{naturemag-doi}
\bibliography{references}

\begin{thebibliography}{10}
\urlstyle{rm}
\expandafter\ifx\csname url\endcsname\relax
  \def\url#1{\texttt{#1}}\fi
\expandafter\ifx\csname urlprefix\endcsname\relax\def\urlprefix{URL }\fi
\expandafter\ifx\csname doiprefix\endcsname\relax\def\doiprefix{DOI: }\fi
\providecommand{\bibinfo}[2]{#2}
\providecommand{\eprint}[2][]{\url{#2}}

\bibitem{kalia2015parkinson}
\bibinfo{author}{Kalia, L.~V.} \& \bibinfo{author}{Lang, A.~E.}
\newblock \bibinfo{journal}{\bibinfo{title}{Parkinson's disease}}.
\newblock {\emph{Lancet}} \textbf{\bibinfo{volume}{386}},
  \bibinfo{pages}{896--912}, \doiprefix\url{10.1016/S0140-6736(14)61393-3}
  (\bibinfo{year}{2015}).

\bibitem{obeso2017past}
\bibinfo{author}{Obeso, J.~A.} \emph{et~al.}
\newblock \bibinfo{journal}{\bibinfo{title}{Past, present, and future of
  {Parkinson's} disease: A special essay on the 200th anniversary of the
  {Shaking Palsy}}}.
\newblock {\emph{Mov. Disord.}}
  \textbf{\bibinfo{volume}{32}}, \bibinfo{pages}{1264--1310},
  \doiprefix\url{10.1002/mds.27115} (\bibinfo{year}{2017}).

\bibitem{dorsey2018global}
\bibinfo{author}{Dorsey, E.~R.} \emph{et~al.}
\newblock \bibinfo{journal}{\bibinfo{title}{Global, regional, and national
  burden of {Parkinson's} disease, 1990--2016: A systematic analysis for the
  {Global Burden of Disease Study 2016}}}.
\newblock {\emph{Lancet Neurol.}}
  \textbf{\bibinfo{volume}{17}}, \bibinfo{pages}{939--953},
  \doiprefix\url{10.1016/S1474-4422(18)30295-3} (\bibinfo{year}{2018}).

\bibitem{benabid2009deep}
\bibinfo{author}{Benabid, A.~L.}, \bibinfo{author}{Chabardes, S.},
  \bibinfo{author}{Mitrofanis, J.} \& \bibinfo{author}{Pollak, P.}
\newblock \bibinfo{journal}{\bibinfo{title}{Deep brain stimulation of the
  subthalamic nucleus for the treatment of {Parkinson's} disease}}.
\newblock {\emph{Lancet Neurol.}}
  \textbf{\bibinfo{volume}{8}}, \bibinfo{pages}{67--81},
  \doiprefix\url{10.1016/S1474-4422(08)70291-6} (\bibinfo{year}{2009}).

\bibitem{chaturvedi2017quantitative}
\bibinfo{author}{Chaturvedi, M.} \emph{et~al.}
\newblock \bibinfo{journal}{\bibinfo{title}{Quantitative {EEG} ({QEEG})
  measures differentiate {Parkinson's} disease ({PD}) patients from healthy
  controls ({HC})}}.
\newblock {\emph{Front. Aging Neurosci.}}
  \textbf{\bibinfo{volume}{9}}, \bibinfo{pages}{3},
  \doiprefix\url{10.3389/fnagi.2017.00003} (\bibinfo{year}{2017}).

\bibitem{oh2020deep}
\bibinfo{author}{Oh, S.~L.} \emph{et~al.}
\newblock \bibinfo{journal}{\bibinfo{title}{A deep learning approach for
  {Parkinson's} disease diagnosis from {EEG} signals}}.
\newblock {\emph{Neural Comput. Appl.}}
  \textbf{\bibinfo{volume}{32}}, \bibinfo{pages}{10927--10933},
  \doiprefix\url{10.1007/s00521-018-3689-5} (\bibinfo{year}{2020}).

\bibitem{abumalloh2024parkinson}
\bibinfo{author}{Abumalloh, R.~A.} \emph{et~al.}
\newblock \bibinfo{journal}{\bibinfo{title}{{Parkinson's} disease diagnosis
  using deep learning: A bibliometric analysis and literature review}}.
\newblock {\emph{Ageing Res. Rev.}}
  \textbf{\bibinfo{volume}{96}}, \bibinfo{pages}{102285},
  \doiprefix\url{10.1016/j.arr.2024.102285} (\bibinfo{year}{2024}).

\bibitem{neves2024parkinson}
\bibinfo{author}{Neves, C.}, \bibinfo{author}{Zeng, Y.} \&
  \bibinfo{author}{Xiao, Y.}
\newblock \bibinfo{title}{{Parkinson's} disease detection from resting state
  {EEG} using multi-head graph structure learning with gradient weighted graph
  attention explanations}.
\newblock In \bibinfo{editor}{Bathula, D.~R.} \emph{et~al.} (eds.)
  \emph{\bibinfo{booktitle}{Machine Learning in Clinical Neuroimaging}}, vol.
  \bibinfo{volume}{15266} of \emph{\bibinfo{series}{Lecture Notes in Computer
  Science}}, \bibinfo{pages}{3--12},
  \doiprefix\url{10.1007/978-3-031-78761-4\_1} (\bibinfo{publisher}{Springer},
  \bibinfo{address}{Cham}, \bibinfo{year}{2024}).

\bibitem{ali2022predictive}
\bibinfo{author}{Ali, A. A.~N.} \emph{et~al.}
\newblock \bibinfo{journal}{\bibinfo{title}{Predictive accuracy of {CNN} for
  cortical oscillatory activity in an acute rat model of parkinsonism}}.
\newblock {\emph{Neural Networks}}
  \textbf{\bibinfo{volume}{146}}, \bibinfo{pages}{334--340},
  \doiprefix\url{10.1016/j.neunet.2021.11.025} (\bibinfo{year}{2022}).

\bibitem{kim2025explainable}
\bibinfo{author}{Kim, J.} \emph{et~al.}
\newblock \bibinfo{title}{Explainable {AI}-driven neural activity analysis in
  {Parkinsonian} rats under electrical stimulation},
  \doiprefix\url{10.48550/arXiv.2502.12471} (\bibinfo{year}{2025}).
\newblock \eprint{2502.12471}.

\bibitem{lachaux2012high}
\bibinfo{author}{Lachaux, J.~P.}, \bibinfo{author}{Axmacher, N.},
  \bibinfo{author}{Mormann, F.}, \bibinfo{author}{Halgren, E.} \&
  \bibinfo{author}{Crone, N.~E.}
\newblock \bibinfo{journal}{\bibinfo{title}{High-frequency neural activity and
  human cognition: Past, present and possible future of intracranial {EEG}
  research}}.
\newblock {\emph{Prog. Neurobiol.}}
  \textbf{\bibinfo{volume}{98}}, \bibinfo{pages}{279--301},
  \doiprefix\url{10.1016/j.pneurobio.2012.06.008} (\bibinfo{year}{2012}).

\bibitem{srinivasan2007eeg}
\bibinfo{author}{Srinivasan, R.}, \bibinfo{author}{Winter, W.~R.},
  \bibinfo{author}{Ding, J.} \& \bibinfo{author}{Nunez, P.~L.}
\newblock \bibinfo{journal}{\bibinfo{title}{{EEG} and {MEG} coherence: Measures
  of functional connectivity at distinct spatial scales of neocortical
  dynamics}}.
\newblock {\emph{J. Neurosci. Methods}}
  \textbf{\bibinfo{volume}{166}}, \bibinfo{pages}{41--52},
  \doiprefix\url{10.1016/j.jneumeth.2007.06.026} (\bibinfo{year}{2007}).

\bibitem{kanth2020electrocorticogram}
\bibinfo{author}{Kanth, S.~T.} \& \bibinfo{author}{Ray, S.}
\newblock \bibinfo{journal}{\bibinfo{title}{Electrocorticogram ({ECoG}) is
  highly informative in primate visual cortex}}.
\newblock {\emph{J. Neurosci.}}
  \textbf{\bibinfo{volume}{40}}, \bibinfo{pages}{2430--2444},
  \doiprefix\url{10.1523/JNEUROSCI.1368-19.2020} (\bibinfo{year}{2020}).

\bibitem{shin2025wireless}
\bibinfo{author}{Shin, H.} \emph{et~al.}
\newblock \bibinfo{journal}{\bibinfo{title}{A wireless cortical surface implant
  for diagnosing and alleviating {Parkinson's} disease symptoms in freely
  moving animals}}.
\newblock {\emph{Adv. Healthc. Mater.}}
  \textbf{\bibinfo{volume}{14}}, \bibinfo{pages}{2405179},
  \doiprefix\url{10.1002/adhm.202405179} (\bibinfo{year}{2025}).

\bibitem{wang2024longitudinal}
\bibinfo{author}{Wang, X.} \emph{et~al.}
\newblock \bibinfo{journal}{\bibinfo{title}{A longitudinal electrophysiological
  and behavior dataset for {PD} rat in response to deep brain stimulation}}.
\newblock {\emph{Sci. Data}}
  \textbf{\bibinfo{volume}{11}}, \bibinfo{pages}{500},
  \doiprefix\url{10.1038/s41597-024-03356-3} (\bibinfo{year}{2024}).

\bibitem{miller2010cortical}
\bibinfo{author}{Miller, K.~J.} \emph{et~al.}
\newblock \bibinfo{journal}{\bibinfo{title}{Cortical activity during motor
  execution, motor imagery, and imagery-based online feedback}}.
\newblock {\emph{Proc. Natl. Acad. Sci. United States Am.}} \textbf{\bibinfo{volume}{107}},
  \bibinfo{pages}{4430--4435}, \doiprefix\url{10.1073/pnas.0913697107}
  (\bibinfo{year}{2010}).

\bibitem{ung2017intracranial}
\bibinfo{author}{Ung, H.} \emph{et~al.}
\newblock \bibinfo{journal}{\bibinfo{title}{Intracranial {EEG} fluctuates over
  months after implanting electrodes in human brain}}.
\newblock {\emph{J. Neural Eng.}}
  \textbf{\bibinfo{volume}{14}}, \bibinfo{pages}{056011},
  \doiprefix\url{10.1088/1741-2552/aa7f40} (\bibinfo{year}{2017}).

\bibitem{ganin2016domain}
\bibinfo{author}{Ganin, Y.} \emph{et~al.}
\newblock \bibinfo{journal}{\bibinfo{title}{Domain-adversarial training of
  neural networks}}.
\newblock {\emph{J. Mach. Learn. Res.}}
  \textbf{\bibinfo{volume}{17}}, \bibinfo{pages}{1--35} (\bibinfo{year}{2016}).

\bibitem{sliwowski2023impact}
\bibinfo{author}{{\'S}liwowski, M.}, \bibinfo{author}{Martin, M.},
  \bibinfo{author}{Souloumiac, A.}, \bibinfo{author}{Blanchart, P.} \&
  \bibinfo{author}{Aksenova, T.}
\newblock \bibinfo{journal}{\bibinfo{title}{Impact of dataset size and
  long-term {ECoG}-based {BCI} usage on deep learning decoders performance}}.
\newblock {\emph{Front. Hum. Neurosci.}}
  \textbf{\bibinfo{volume}{17}}, \bibinfo{pages}{1111645},
  \doiprefix\url{10.3389/fnhum.2023.1111645} (\bibinfo{year}{2023}).

\bibitem{haufe2018elucidating}
\bibinfo{author}{Haufe, S.} \emph{et~al.}
\newblock \bibinfo{journal}{\bibinfo{title}{Elucidating relations between
  {fMRI}, {ECoG}, and {EEG} through a common natural stimulus}}.
\newblock {\emph{NeuroImage}} \textbf{\bibinfo{volume}{179}},
  \bibinfo{pages}{79--91}, \doiprefix\url{10.1016/j.neuroimage.2018.06.016}
  (\bibinfo{year}{2018}).

\bibitem{memar2025rise}
\bibinfo{author}{Memar, M.~O.}, \bibinfo{author}{Ziaei, N.},
  \bibinfo{author}{Nazari, B.} \& \bibinfo{author}{Yousefi, A.}
\newblock \bibinfo{title}{{RISE-iEEG}: Robust to inter-subject electrodes
  implantation variability {iEEG} classifier}.
\newblock In \emph{\bibinfo{booktitle}{2025 47th Annual International
  Conference of the IEEE Engineering in Medicine and Biology Society (EMBC)}},
  \bibinfo{pages}{1--7}, \doiprefix\url{10.1109/EMBC58623.2025.11252788}
  (\bibinfo{publisher}{IEEE}, \bibinfo{year}{2025}).

\bibitem{dayanik2021disentangling}
\bibinfo{author}{Dayanik, E.} \& \bibinfo{author}{Pad{\'o}, S.}
\newblock \bibinfo{title}{Disentangling document topic and author gender in
  multiple languages: Lessons for adversarial debiasing}.
\newblock In \emph{\bibinfo{booktitle}{Proceedings of the Eleventh Workshop on
  Computational Approaches to Subjectivity, Sentiment and Social Media
  Analysis}}, \bibinfo{pages}{50--61} (\bibinfo{publisher}{Association for
  Computational Linguistics}, \bibinfo{address}{Online}, \bibinfo{year}{2021}).

\bibitem{mullen2015real}
\bibinfo{author}{Mullen, T.~R.} \emph{et~al.}
\newblock \bibinfo{journal}{\bibinfo{title}{Real-time neuroimaging and
  cognitive monitoring using wearable dry {EEG}}}.
\newblock {\emph{IEEE Transactions on Biomed. Eng.}}
  \textbf{\bibinfo{volume}{62}}, \bibinfo{pages}{2553--2567},
  \doiprefix\url{10.1109/TBME.2015.2481482} (\bibinfo{year}{2015}).

\bibitem{leergaard2000somatotopic}
\bibinfo{author}{Leergaard, T.~B.}, \bibinfo{author}{Alloway, K.~D.},
  \bibinfo{author}{Mutic, J.~J.} \& \bibinfo{author}{Bjaalie, J.~G.}
\newblock \bibinfo{journal}{\bibinfo{title}{Three-dimensional topography of
  corticopontine projections from rat barrel cortex: Correlations with
  corticostriatal organization}}.
\newblock {\emph{J. Neurosci.}}
  \textbf{\bibinfo{volume}{20}}, \bibinfo{pages}{8474--8484},
  \doiprefix\url{10.1523/JNEUROSCI.20-22-08474.2000} (\bibinfo{year}{2000}).

\bibitem{penfield1937somatic}
\bibinfo{author}{Penfield, W.} \& \bibinfo{author}{Boldrey, E.}
\newblock \bibinfo{journal}{\bibinfo{title}{Somatic motor and sensory
  representation in the cerebral cortex of man as studied by electrical
  stimulation}}.
\newblock {\emph{Brain}} \textbf{\bibinfo{volume}{60}},
  \bibinfo{pages}{389--443}, \doiprefix\url{10.1093/brain/60.4.389}
  (\bibinfo{year}{1937}).

\bibitem{tang2025sdc}
\bibinfo{author}{Tang, J.} \emph{et~al.}
\newblock \bibinfo{title}{{SDC-Net}: A domain adaptation framework with
  semantic-dynamic consistency for cross-subject {EEG} emotion recognition},
  \doiprefix\url{10.48550/arXiv.2507.17524} (\bibinfo{year}{2025}).
\newblock \eprint{2507.17524}.

\bibitem{saeed2021learning}
\bibinfo{author}{Saeed, A.}, \bibinfo{author}{Grangier, D.},
  \bibinfo{author}{Pietquin, O.} \& \bibinfo{author}{Zeghidour, N.}
\newblock \bibinfo{title}{Learning from heterogeneous {EEG} signals with
  differentiable channel reordering}.
\newblock In \emph{\bibinfo{booktitle}{ICASSP 2021--2021 IEEE International
  Conference on Acoustics, Speech and Signal Processing (ICASSP)}},
  \bibinfo{pages}{1255--1259}, \doiprefix\url{10.1109/ICASSP39728.2021.9413712}
  (\bibinfo{publisher}{IEEE}, \bibinfo{year}{2021}).

\bibitem{agrawal2025quantitative}
\bibinfo{author}{Agrawal, S.} \& \bibinfo{author}{Sahu, S.~P.}
\newblock \bibinfo{title}{Quantitative {EEG}-based regional spectral analysis
  for {Parkinson's} disease detection}.
\newblock In \emph{\bibinfo{booktitle}{2025 IEEE International Conference on
  Interdisciplinary Approaches in Technology and Management for Social
  Innovation}}, vol.~\bibinfo{volume}{3}, \bibinfo{pages}{1--6},
  \doiprefix\url{10.1109/IATMSI64286.2025.10985366} (\bibinfo{publisher}{IEEE},
  \bibinfo{year}{2025}).

\bibitem{jeong2016wavelet}
\bibinfo{author}{Jeong, D.-H.}, \bibinfo{author}{Kim, Y.-D.},
  \bibinfo{author}{Song, I.-U.}, \bibinfo{author}{Chung, Y.-A.} \&
  \bibinfo{author}{Jeong, J.}
\newblock \bibinfo{journal}{\bibinfo{title}{Wavelet energy and wavelet
  coherence as {EEG} biomarkers for the diagnosis of {Parkinson's}
  disease-related dementia and {Alzheimer's} disease}}.
\newblock {\emph{Entropy}} \textbf{\bibinfo{volume}{18}},
  \bibinfo{pages}{8}, \doiprefix\url{10.3390/e18010008} (\bibinfo{year}{2016}).

\bibitem{obayya2023novel}
\bibinfo{author}{Obayya, M.} \emph{et~al.}
\newblock \bibinfo{journal}{\bibinfo{title}{A novel automated {Parkinson's}
  disease identification approach using deep learning and {EEG}}}.
\newblock {\emph{PeerJ Comput. Sci.}}
  \textbf{\bibinfo{volume}{9}}, \bibinfo{pages}{e1663},
  \doiprefix\url{10.7717/peerj-cs.1663} (\bibinfo{year}{2023}).

\bibitem{schlogl1996single}
\bibinfo{author}{Schl{\"o}gl, A.}, \bibinfo{author}{Pfurtscheller, G.} \&
  \bibinfo{author}{Schack, B.}
\newblock \bibinfo{title}{Single-trial {EEG} analysis using an adaptive
  autoregressive model}.
\newblock In \emph{\bibinfo{booktitle}{Proceedings of the Fourth International
  Symposium on Central Nervous Monitoring}} (\bibinfo{address}{Gmunden,
  Austria}, \bibinfo{year}{1996}).
\newblock \bibinfo{note}{September 5--7, 1996}.

\bibitem{shah2020dynamical}
\bibinfo{author}{Shah, S. A.~A.}, \bibinfo{author}{Zhang, L.} \&
  \bibinfo{author}{Bais, A.}
\newblock \bibinfo{journal}{\bibinfo{title}{Dynamical system based compact deep
  hybrid network for classification of {Parkinson} disease related {EEG}
  signals}}.
\newblock {\emph{Neural Networks}}
  \textbf{\bibinfo{volume}{130}}, \bibinfo{pages}{75--84},
  \doiprefix\url{10.1016/j.neunet.2020.06.018} (\bibinfo{year}{2020}).

\bibitem{bore2020prediction}
\bibinfo{author}{Bore, J.~C.} \emph{et~al.}
\newblock \bibinfo{journal}{\bibinfo{title}{Prediction of mild parkinsonism
  revealed by neural oscillatory changes and machine learning}}.
\newblock {\emph{J. Neurophysiol.}}
  \textbf{\bibinfo{volume}{124}}, \bibinfo{pages}{1698--1705},
  \doiprefix\url{10.1152/jn.00534.2020} (\bibinfo{year}{2020}).

\bibitem{lam2024self}
\bibinfo{author}{Lam, V.}, \bibinfo{author}{Oliugbo, C.},
  \bibinfo{author}{Parida, A.}, \bibinfo{author}{Linguraru, M.~G.} \&
  \bibinfo{author}{Anwar, S.~M.}
\newblock \bibinfo{title}{Self-supervised learning for seizure classification
  using {ECoG} spectrograms}.
\newblock In \emph{\bibinfo{booktitle}{Medical Imaging 2024: Computer-Aided
  Diagnosis}}, vol. \bibinfo{volume}{12927} of
  \emph{\bibinfo{series}{Proceedings of SPIE}}, \bibinfo{pages}{129272J},
  \doiprefix\url{10.1117/12.3007431} (\bibinfo{publisher}{SPIE},
  \bibinfo{year}{2024}).

\bibitem{ji2024bi}
\bibinfo{author}{Ji, C.}
\newblock \bibinfo{title}{Bi-band {ECoGNet} for {ECoG} decoding on
  classification task}, \doiprefix\url{10.48550/arXiv.2412.00378}
  (\bibinfo{year}{2024}).
\newblock \eprint{2412.00378}.

\bibitem{cui2022cluster}
\bibinfo{author}{Cui, X.}, \bibinfo{author}{Cao, J.}, \bibinfo{author}{Lai,
  X.}, \bibinfo{author}{Jiang, T.} \& \bibinfo{author}{Gao, F.}
\newblock \bibinfo{journal}{\bibinfo{title}{Cluster embedding
  joint-probability-discrepancy transfer for cross-subject seizure detection}}.
\newblock {\emph{IEEE Transactions on Neural Syst. Rehabil. Eng.}} \textbf{\bibinfo{volume}{31}},
  \bibinfo{pages}{593--605}, \doiprefix\url{10.1109/TNSRE.2022.3229066}
  (\bibinfo{year}{2023}).

\bibitem{li2018domain}
\bibinfo{author}{Li, H.}, \bibinfo{author}{Pan, S.~J.}, \bibinfo{author}{Wang,
  S.} \& \bibinfo{author}{Kot, A.~C.}
\newblock \bibinfo{title}{Domain generalization with adversarial feature
  learning}.
\newblock In \emph{\bibinfo{booktitle}{Proceedings of the IEEE/CVF Conference
  on Computer Vision and Pattern Recognition (CVPR)}},
  \bibinfo{pages}{5400--5409}, \doiprefix\url{10.1109/CVPR.2018.00566}
  (\bibinfo{publisher}{IEEE}, \bibinfo{year}{2018}).

\bibitem{dayal2023madg}
\bibinfo{author}{Dayal, A.} \emph{et~al.}
\newblock \bibinfo{title}{{MADG}: Margin-based adversarial learning for domain
  generalization}.
\newblock In \emph{\bibinfo{booktitle}{Advances in Neural Information
  Processing Systems}}, vol.~\bibinfo{volume}{36},
  \bibinfo{pages}{58938--58952} (\bibinfo{year}{2023}).

\bibitem{shankar2018generalizing}
\bibinfo{author}{Shankar, S.} \emph{et~al.}
\newblock \bibinfo{title}{Generalizing across domains via cross-gradient
  training}.
\newblock In \emph{\bibinfo{booktitle}{\aqrevision{Proceedings of the}
  International Conference on Learning Representations (ICLR)}}
  (\bibinfo{year}{2018}).

\bibitem{muandet2013domain}
\bibinfo{author}{Muandet, K.}, \bibinfo{author}{Balduzzi, D.} \&
  \bibinfo{author}{Sch{\"o}lkopf, B.}
\newblock \bibinfo{title}{Domain generalization via invariant feature
  representation}.
\newblock In \emph{\bibinfo{booktitle}{Proceedings of the 30th International
  Conference on Machine Learning}}, vol.~\bibinfo{volume}{28} of
  \emph{\bibinfo{series}{Proceedings of Machine Learning Research}},
  \bibinfo{pages}{10--18} (\bibinfo{publisher}{PMLR}, \bibinfo{year}{2013}).

\bibitem{arjovsky2020invariant}
\bibinfo{author}{Arjovsky, M.}, \bibinfo{author}{Bottou, L.},
  \bibinfo{author}{Gulrajani, I.} \& \bibinfo{author}{Lopez-Paz, D.}
\newblock \bibinfo{title}{Invariant risk minimization},
  \doiprefix\url{10.48550/arXiv.1907.02893} (\bibinfo{year}{2019}).
\newblock \eprint{1907.02893}.

\bibitem{sagawa2019distributionally}
\bibinfo{author}{Sagawa, S.}, \bibinfo{author}{Koh, P.~W.},
  \bibinfo{author}{Hashimoto, T.~B.} \& \bibinfo{author}{Liang, P.}
\newblock \bibinfo{title}{Distributionally robust neural networks for group
  shifts: On the importance of regularization for worst-case generalization}.
\newblock In \emph{\bibinfo{booktitle}{\aqrevision{Proceedings of the}
  International Conference on Learning Representations (ICLR)}}
  (\bibinfo{year}{2020}).

\bibitem{yang2020towards}
\bibinfo{author}{Yang, J.}, \bibinfo{author}{Zou, H.}, \bibinfo{author}{Zhou,
  Y.} \& \bibinfo{author}{Xie, L.}
\newblock \bibinfo{title}{Towards stable and comprehensive domain alignment:
  Max-margin domain-adversarial training},
  \doiprefix\url{10.48550/arXiv.2003.13249} (\bibinfo{year}{2020}).
\newblock \eprint{2003.13249}.

\bibitem{ma2019reducing}
\bibinfo{author}{Ma, B.-Q.}, \bibinfo{author}{Li, H.}, \bibinfo{author}{Zheng,
  W.-L.} \& \bibinfo{author}{Lu, B.-L.}
\newblock \bibinfo{title}{Reducing the subject variability of {EEG} signals
  with adversarial domain generalization}.
\newblock In \bibinfo{editor}{Gedeon, T.}, \bibinfo{editor}{Wong, K.~W.} \&
  \bibinfo{editor}{Lee, M.} (eds.) \emph{\bibinfo{booktitle}{Neural Information
  Processing}}, vol. \bibinfo{volume}{11953} of \emph{\bibinfo{series}{Lecture
  Notes in Computer Science}}, \bibinfo{pages}{30--42},
  \doiprefix\url{10.1007/978-3-030-36708-4\_3} (\bibinfo{publisher}{Springer},
  \bibinfo{address}{Cham}, \bibinfo{year}{2019}).

\bibitem{wang2024dmmr}
\bibinfo{author}{Wang, Y.}, \bibinfo{author}{Zhang, B.} \&
  \bibinfo{author}{Tang, Y.}
\newblock \bibinfo{journal}{\bibinfo{title}{{DMMR}: Cross-subject domain
  generalization for {EEG}-based emotion recognition via denoising mixed mutual
  reconstruction}}.
\newblock {\emph{Proc. AAAI Conf. on Artif. Intell.}} \textbf{\bibinfo{volume}{38}}, \bibinfo{pages}{628--636},
  \doiprefix\url{10.1609/aaai.v38i1.27819} (\bibinfo{year}{2024}).

\bibitem{tao2024local}
\bibinfo{author}{Tao, J.}, \bibinfo{author}{Dan, Y.} \& \bibinfo{author}{Zhou,
  D.}
\newblock \bibinfo{journal}{\bibinfo{title}{Local domain generalization with
  low-rank constraint for {EEG}-based emotion recognition}}.
\newblock {\emph{Front. Neurosci.}}
  \textbf{\bibinfo{volume}{17}}, \bibinfo{pages}{1213099},
  \doiprefix\url{10.3389/fnins.2023.1213099} (\bibinfo{year}{2023}).

\bibitem{zhang2018mixup}
\bibinfo{author}{Zhang, H.}, \bibinfo{author}{Cisse, M.},
  \bibinfo{author}{Dauphin, Y.~N.} \& \bibinfo{author}{Lopez-Paz, D.}
\newblock \bibinfo{title}{mixup: Beyond empirical risk minimization}.
\newblock In \emph{\bibinfo{booktitle}{\aqrevision{Proceedings of the}
  International Conference on Learning Representations (ICLR)}}
  (\bibinfo{year}{2018}).

\bibitem{yun2019cutmix}
\bibinfo{author}{Yun, S.} \emph{et~al.}
\newblock \bibinfo{title}{{CutMix}: Regularization strategy to train strong
  classifiers with localizable features}.
\newblock In \emph{\bibinfo{booktitle}{Proceedings of the IEEE/CVF
  International Conference on Computer Vision (ICCV)}},
  \bibinfo{pages}{6022--6031}, \doiprefix\url{10.1109/ICCV.2019.00612}
  (\bibinfo{publisher}{IEEE}, \bibinfo{year}{2019}).

\bibitem{song2023eeg}
\bibinfo{author}{Song, Y.}, \bibinfo{author}{Zheng, Q.}, \bibinfo{author}{Liu,
  B.} \& \bibinfo{author}{Gao, X.}
\newblock \bibinfo{journal}{\bibinfo{title}{{EEG} conformer: Convolutional
  transformer for {EEG} decoding and visualization}}.
\newblock {\emph{IEEE Transactions on Neural Syst. Rehabil. Eng.}} \textbf{\bibinfo{volume}{31}},
  \bibinfo{pages}{710--719}, \doiprefix\url{10.1109/TNSRE.2022.3230250}
  (\bibinfo{year}{2023}).

\bibitem{rommel2022cadda}
\bibinfo{author}{Rommel, C.}, \bibinfo{author}{Moreau, T.},
  \bibinfo{author}{Paillard, J.} \& \bibinfo{author}{Gramfort, A.}
\newblock \bibinfo{title}{{CADDA}: Class-wise automatic differentiable data
  augmentation for {EEG} signals}.
\newblock In \emph{\bibinfo{booktitle}{\aqrevision{Proceedings of the}
  International Conference on Learning Representations (ICLR)}}
  (\bibinfo{year}{2022}).

\bibitem{mohsenvand2020contrastive}
\bibinfo{author}{Mohsenvand, M.~N.}, \bibinfo{author}{Izadi, M.~R.} \&
  \bibinfo{author}{Maes, P.}
\newblock \bibinfo{title}{Contrastive representation learning for
  electroencephalogram classification}.
\newblock In \emph{\bibinfo{booktitle}{Proceedings of the Machine Learning for
  Health NeurIPS Workshop}}, vol. \bibinfo{volume}{136} of
  \emph{\bibinfo{series}{Proceedings of Machine Learning Research}},
  \bibinfo{pages}{238--253} (\bibinfo{publisher}{PMLR}, \bibinfo{year}{2020}).

\bibitem{lawhern2018eegnet}
\bibinfo{author}{Lawhern, V.~J.} \emph{et~al.}
\newblock \bibinfo{journal}{\bibinfo{title}{{EEGNet}: A compact convolutional
  neural network for {EEG}-based brain-computer interfaces}}.
\newblock {\emph{J. Neural Eng.}}
  \textbf{\bibinfo{volume}{15}}, \bibinfo{pages}{056013},
  \doiprefix\url{10.1088/1741-2552/aace8c} (\bibinfo{year}{2018}).

\bibitem{cui2023towards}
\bibinfo{author}{Cui, J.}, \bibinfo{author}{Yuan, L.}, \bibinfo{author}{Wang,
  Z.}, \bibinfo{author}{Li, R.} \& \bibinfo{author}{Jiang, T.}
\newblock \bibinfo{journal}{\bibinfo{title}{Towards best practice of
  interpreting deep learning models for {EEG}-based brain computer
  interfaces}}.
\newblock {\emph{Front. Comput. Neurosci.}}
  \textbf{\bibinfo{volume}{17}}, \bibinfo{pages}{1232925},
  \doiprefix\url{10.3389/fncom.2023.1232925} (\bibinfo{year}{2023}).

\bibitem{chang2018evaluation}
\bibinfo{author}{Chang, C.-Y.}, \bibinfo{author}{Hsu, S.-H.},
  \bibinfo{author}{Pion-Tonachini, L.} \& \bibinfo{author}{Jung, T.-P.}
\newblock \bibinfo{title}{Evaluation of artifact subspace reconstruction for
  automatic {EEG} artifact removal}.
\newblock In \emph{\bibinfo{booktitle}{2018 40th Annual International
  Conference of the IEEE Engineering in Medicine and Biology Society (EMBC)}},
  \bibinfo{pages}{1242--1245}, \doiprefix\url{10.1109/EMBC.2018.8512547}
  (\bibinfo{publisher}{IEEE}, \bibinfo{year}{2018}).

\bibitem{huang2023discrepancy}
\bibinfo{author}{Huang, G.} \emph{et~al.}
\newblock \bibinfo{journal}{\bibinfo{title}{Discrepancy between inter- and
  intra-subject variability in {EEG}-based motor imagery brain-computer
  interface: Evidence from multiple perspectives}}.
\newblock {\emph{Front. Neurosci.}}
  \textbf{\bibinfo{volume}{17}}, \bibinfo{pages}{1122661},
  \doiprefix\url{10.3389/fnins.2023.1122661} (\bibinfo{year}{2023}).

\bibitem{kim2019learning}
\bibinfo{author}{Kim, B.}, \bibinfo{author}{Kim, H.}, \bibinfo{author}{Kim,
  K.}, \bibinfo{author}{Kim, S.} \& \bibinfo{author}{Kim, J.}
\newblock \bibinfo{title}{Learning not to learn: Training deep neural networks
  with biased data}.
\newblock In \emph{\bibinfo{booktitle}{Proceedings of the IEEE/CVF Conference
  on Computer Vision and Pattern Recognition (CVPR)}},
  \bibinfo{pages}{9004--9012}, \doiprefix\url{10.1109/CVPR.2019.00922}
  (\bibinfo{publisher}{IEEE}, \bibinfo{year}{2019}).

\bibitem{kyllonen2024unsupervised}
\bibinfo{author}{Kyll{\"o}nen, M.}
\newblock \emph{\bibinfo{title}{Unsupervised representation learning
  visualization for brain activity dynamics in {REM}-sleep}}.
\newblock Master's thesis, \bibinfo{school}{Aalto University},
  \bibinfo{address}{Espoo, Finland} (\bibinfo{year}{2024}).

\bibitem{cavanagh2018diminished}
\bibinfo{author}{Cavanagh, J.~F.}, \bibinfo{author}{Kumar, P.},
  \bibinfo{author}{Mueller, A.~A.}, \bibinfo{author}{Richardson, S.~P.} \&
  \bibinfo{author}{Mueen, A.}
\newblock \bibinfo{journal}{\bibinfo{title}{Diminished {EEG} habituation to
  novel events effectively classifies {Parkinson's} patients}}.
\newblock {\emph{Clin. Neurophysiol.}}
  \textbf{\bibinfo{volume}{129}}, \bibinfo{pages}{409--418},
  \doiprefix\url{10.1016/j.clinph.2017.11.023} (\bibinfo{year}{2018}).

\bibitem{singh2020frontal}
\bibinfo{author}{Singh, A.} \emph{et~al.}
\newblock \bibinfo{journal}{\bibinfo{title}{Frontal theta and beta oscillations
  during lower-limb movement in {Parkinson's} disease}}.
\newblock {\emph{Clin. Neurophysiol.}}
  \textbf{\bibinfo{volume}{131}}, \bibinfo{pages}{694--702},
  \doiprefix\url{10.1016/j.clinph.2019.12.399} (\bibinfo{year}{2020}).

\bibitem{sugden2023generalizable}
\bibinfo{author}{Sugden, R.~J.} \& \bibinfo{author}{Diamandis, P.}
\newblock \bibinfo{journal}{\bibinfo{title}{Generalizable
  electroencephalographic classification of {Parkinson's} disease using deep
  learning}}.
\newblock {\emph{Informatics Medicine Unlocked}}
  \textbf{\bibinfo{volume}{42}}, \bibinfo{pages}{101352},
  \doiprefix\url{10.1016/j.imu.2023.101352} (\bibinfo{year}{2023}).

\bibitem{karakas2023distinguishing}
\bibinfo{author}{Karaka{\c{s}}, M.~F.} \& \bibinfo{author}{Latifo{\u{g}}lu, F.}
\newblock \bibinfo{journal}{\bibinfo{title}{Distinguishing {Parkinson's}
  disease with {GLCM} features from the hankelization of {EEG} signals}}.
\newblock {\emph{Diagnostics}} \textbf{\bibinfo{volume}{13}},
  \bibinfo{pages}{1769}, \doiprefix\url{10.3390/diagnostics13101769}
  (\bibinfo{year}{2023}).

\bibitem{wang2025dbconformer}
\bibinfo{author}{Wang, Z.} \emph{et~al.}
\newblock \bibinfo{journal}{\bibinfo{title}{{DBConformer}: Dual-branch
  convolutional {Transformer} for {EEG} decoding}}.
\newblock {\emph{IEEE J. Biomed. Heal. Informatics}} \textbf{\bibinfo{volume}{30}}, \bibinfo{pages}{4134--4147},
  \doiprefix\url{10.1109/JBHI.2025.3622725} (\bibinfo{year}{2026}).

\bibitem{altaheri2025temporal}
\bibinfo{author}{Altaheri, H.}, \bibinfo{author}{Karray, F.} \&
  \bibinfo{author}{Karimi, A.~H.}
\newblock \bibinfo{journal}{\bibinfo{title}{Temporal convolutional transformer
  for {EEG} based motor imagery decoding}}.
\newblock {\emph{Sci. Reports}}
  \textbf{\bibinfo{volume}{15}}, \bibinfo{pages}{32959},
  \doiprefix\url{10.1038/s41598-025-16219-7} (\bibinfo{year}{2025}).

\bibitem{zhong2024eegdg}
\bibinfo{author}{Zhong, X.-C.} \emph{et~al.}
\newblock \bibinfo{journal}{\bibinfo{title}{{EEG-DG}: A multi-source domain
  generalization framework for motor imagery {EEG} classification}}.
\newblock {\emph{IEEE J. Biomed. Heal. Informatics}} \textbf{\bibinfo{volume}{29}}, \bibinfo{pages}{2484--2495},
  \doiprefix\url{10.1109/JBHI.2024.3431230} (\bibinfo{year}{2025}).

\bibitem{liang2025secnet}
\bibinfo{author}{Liang, W.} \emph{et~al.}
\newblock \bibinfo{journal}{\bibinfo{title}{{SecNet}: A second order neural
  network for {MI-EEG}}}.
\newblock {\emph{Inf. Process. Manag.}}
  \textbf{\bibinfo{volume}{62}}, \bibinfo{pages}{104012},
  \doiprefix\url{10.1016/j.ipm.2024.104012} (\bibinfo{year}{2025}).

\bibitem{liu2023fbmsnet}
\bibinfo{author}{Liu, K.}, \bibinfo{author}{Yang, M.}, \bibinfo{author}{Yu,
  Z.}, \bibinfo{author}{Wang, G.} \& \bibinfo{author}{Wu, W.}
\newblock \bibinfo{journal}{\bibinfo{title}{{FBMSNet}: A filter-bank
  multi-scale convolutional neural network for {EEG}-based motor imagery
  decoding}}.
\newblock {\emph{IEEE Trans. Biomed. Eng.}}
  \textbf{\bibinfo{volume}{70}}, \bibinfo{pages}{436--445},
  \doiprefix\url{10.1109/TBME.2022.3193277} (\bibinfo{year}{2023}).

\end{thebibliography}
\end{document}